\documentclass[sigconf]{acmart}

\usepackage{xcolor}
\usepackage{xspace}
\usepackage{cleveref}
\usepackage{subcaption}
\usepackage{booktabs}
\usepackage{pifont}
\usepackage{multirow}
\usepackage{gensymb}
\usepackage{float}

\AtBeginDocument{%
  }

\copyrightyear{2026}
\acmYear{2026}
\setcopyright{cc}
\setcctype{by}
\acmConference[CHI '26]{Proceedings of the 2026 CHI Conference on Human Factors in Computing Systems}{April 13--17, 2026}{Barcelona, Spain}
\acmBooktitle{Proceedings of the 2026 CHI Conference on Human Factors in Computing Systems (CHI '26), April 13--17, 2026, Barcelona, Spain}
\acmDOI{10.1145/3772318.3790493}
\acmISBN{979-8-4007-2278-3/2026/04}

\begin{document}

\title{DeltaDorsal: Enhancing Hand Pose Estimation with Dorsal Features in Egocentric Views}

\author{William Huang}
\orcid{0000-0001-7651-2190}
\affiliation{%
  \institution{University of California, Los Angeles}
  \city{California}
  \country{USA}
}
\email{william.huang@ucla.edu}

\author{Siyou Pei}
\orcid{0000-0003-3802-8298}
\affiliation{%
  \institution{University of California, Los Angeles}
  \city{California}
  \country{USA}
}
\email{sypei@ucla.edu}

\author{Leyi Zou}
\orcid{0009-0000-7689-0036}
\affiliation{%
  \institution{University of California, Los Angeles}
  \city{California}
  \country{USA}
}
\email{zelozou@ucla.edu}

\author{Eric J. Gonzalez}
\orcid{0000-0002-2846-7687}
\affiliation{%
  \institution{Google}
  \city{Seattle}
  \country{USA}
}
\email{ejgonz@google.com}

\author{Ishan Chatterjee}
\orcid{0000-0002-2123-6392}
\affiliation{%
  \institution{Google}
  \city{Seattle}
  \country{USA}
}
\email{ishanc@google.com}

\author{Yang Zhang}
\orcid{0000-0003-2472-6968}
\affiliation{%
  \institution{University of California, Los Angeles}
  \city{Los Angeles}
  \state{CA}
  \country{USA}
}
\email{yangzhang@ucla.edu}

\renewcommand{\shortauthors}{Huang et al.}

\def\systemname {DeltaDorsal\xspace}
\def\eg {\textit{e.g.}\xspace}
\def\ie {\textit{i.e.}\xspace}

\newcommand{\changed}[1]{\textcolor{black}{#1}}

\begin{abstract}

The proliferation of XR devices has made egocentric hand pose estimation a vital task, yet this perspective is inherently challenged by frequent finger occlusions. To address this, we propose a novel approach that leverages the rich information in dorsal hand skin deformation, unlocked by recent advances in dense visual featurizers. We introduce a dual-stream delta encoder that learns pose by contrasting features from a dynamic hand with a baseline relaxed position. Our evaluation demonstrates that, using only cropped dorsal images, our method reduces the Mean Per Joint Angle Error (MPJAE) by \changed{18\% in self-occluded scenarios (fingers $\geq50\%$ occluded)} compared to state-of-the-art techniques that depend on the whole hand's geometry and large model backbones. Consequently, our method not only enhances the reliability of downstream tasks like index finger pinch and tap estimation in occluded scenarios but also unlocks new interaction paradigms, such as detecting isometric force for a surface ``click'' without visible movement while minimizing model size.

\end{abstract}



\begin{teaserfigure}
  \includegraphics[width=\textwidth]{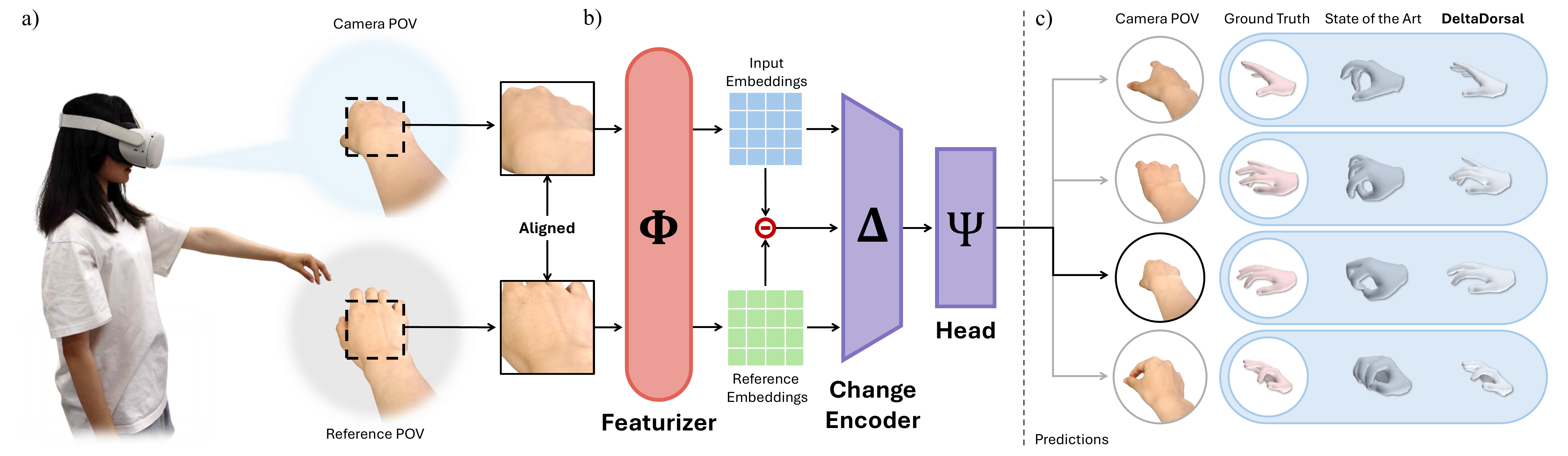}
  \caption{Egocentric hand pose estimation is a challenging problem due to frequent occlusion of the fingers (a). Leveraging recent advances in dense visual featurization, \systemname enables 3D hand pose estimation using purely visual signals from the dorsum of the hand (b). By comparing visual features from a neutral and current hand pose, \systemname isolates skin deformation and uses these signals to predict hand posture, outperforming ``State of the Art'' (HaMeR \cite{pavlakosReconstructingHands3D2023}) models that rely on geometric representations of the full hand (c).}
  \label{fig:teaser}
  \Description{
  Visualization of the system in action. On the left is someone in a VR headset performing a pinch with a popup showing the headset cameras view of the hand with the fingers occluded by the palm. On the bottom is a reference picture of the hand in a neutral position (a). Dorsal features from both hands are cut out and aligned before being fed into a featurizer which creates image embeddings for each, subtracts them, and feeds all outputs into a change encoder and regression head (b). On the right are images of different camera captures with ground truth. From top down, for an open hand the state of the art (SOTA) predicts a circle gesture while DeltaDorsal predicts a grasp. For ring finger pinch, the SOTA predicts all fingers curled in while DeltaDorsal predicts only the ring finger pinching. For an index tap, SOTA predicts a grasp while DeltaDorsal predicts an index tap. For an index pinch, SOTA predicts a near fist while DeltaDorsal predicts a pinch with slightly more distance between the index and thumb. 
  }
\end{teaserfigure}

% \received{11 September 2025}

\maketitle

\section{Introduction}

% 1. Establish importance of hand tracking in HCI. Introduce the problem of self-occlusion
Hand tracking is a cornerstone of modern human–computer interaction \cite{chakravarthiComprehensiveReviewLeap2023, ungureanuHoloLens2Research2020, Zhang2020MediaPipe}, enabling intuitive gesture-based control in Extended Reality (XR), wearables \cite{yeoOpisthenarHandPoses2019}, and other interactive systems \cite{peiHandInterfacesUsing2022, mollynEgoTouchOnBodyTouch2024}. As these technologies expand, the need for accurate and robust hand pose estimation continues to grow, especially from an egocentric point-of-view. However, a key challenge for egocentric hand pose estimation remains: self-occulusion. From this viewpoint, the palm often partially or fully obscures the very fingers being tracked. Our analysis of prominent egocentric hand datasets \cite{Ohkawa2023AssemblyHands, graumanEgoExo4DUnderstandingSkilled2024, fanARCTICDatasetDexterous2023, Kwon2021H2O} reveals that a finger is fully occluded in nearly a fifth of all scenarios, with partial occlusions occurring even more frequently.

Yet, most existing approaches rely solely on the overall hand silhouette or coarse geometric representations of the hand where pose information is preserved only in features like edges and smoothed-out pixel blocks with detailed skin deformation often treated as noise and discarded \cite{ge3DHandShape2019, Zimmermann2017Learning, Mueller2017RealTime, muellerRealtimePoseShape2019}. Our findings show that occlusion in egocentric data poses a significant challenge for these state-of-the-art (SOTA) models, affecting both general-purpose models like HaMeR \cite{pavlakosReconstructingHands3D2023} and even those specifically designed to address hand-object occlusions, such as HandOccNet \cite{parkHandOccNetOcclusionRobust3D2022}. 
This limitation prevents free-hand interaction from achieving its potential to replace conventional input methods in spatial computing scenarios that demand robustness in complex, dynamic real-world settings.

To address these shortcomings, we investigate a rich, untapped source of information that remains visible even during occlusion: dorsal skin deformation. Human skin continuously deforms with tendon and muscle activity, especially in the hands \cite{zhaiInfluenceGraspingPostures2023, chenVivoPanoramicHuman2020, caravaggiAccuracyCorrelationSkinmarker2021, rupaniLocalPosturalChanges2025}. For example, the skin around the knuckles stretches taut when making a fist, while tendons protrude when extending the fingers. These fine-grained deformations encode valuable cues about hand pose and applied force. Unlike silhouette information, these cues remain observable even when the fingers themselves are occluded. Despite this potential, dorsal signals are rarely used. The limited prior works that do use dorsal deformation mount a camera rigidly to the wrist to have a fixed coordinate relation with relation to the back-of-the-hand, limiting wrist motion and adding an additional sensor mounting point for use with egocentric XR systems \cite{wuBackHandPose3DHand2020, yeoOpisthenarHandPoses2019}.

%Prior systems such as DorsalNet \cite{wuBackHandPose3DHand2020} and Opisthenar \cite{yeoOpisthenarHandPoses2019} rely on wrist-mounted cameras and motion images, requiring extra sensors, potentially degrading in unstable video feeds, and constraining wrist movement to within the camera's field-of-view.
%On the other hand, more general hand pose estimation models commonly implemented in XR applications largely ignore dorsal features, often opting to downscale images to use specific computer vision architectures over preserving fine-grain dorsal detail \cite{Zhang2020MediaPipe, parkHandOccNetOcclusionRobust3D2022, pavlakosReconstructingHands3D2023}. 

In this work, we introduce \systemname, the first end-to-end system to use dorsal skin features for egocentric hand pose estimation. Leveraging recent advances in dense visual featurizers \cite{simeoniDINOv32025}, our method captures and amplifies subtle dorsal skin deformations relative to a neutral hand pose. By analyzing the feature differences from vision backbones that capture both visual and semantic information, we isolate these deformation signals, filter noise, and make minute dorsal cues accessible without requiring explicit hand localization via wrist-mounted cameras or temporally-bound motion images. Our system demonstrates similar or better performance to compared to SotA models with a smaller model size and more consistent training.

%This enables dorsal information to be captured using standard egocentric cameras already embedded in XR headsets.

% We motivate this approach to hand pose estimation by analyzing a key limitation of existing hand pose estimation models: self-occlusion. We assess the impacts of self-occlusion on existing model performance and quantify the availability of dorsal features to motivate our work by analyzing existing egocentric hand datasets \cite{Ohkawa2023AssemblyHands, graumanEgoExo4DUnderstandingSkilled2024, fanARCTICDatasetDexterous2023, Kwon2021H2O} and two state-of-the-art (SOTA) hand pose estimation models: HaMeR \cite{pavlakosReconstructingHands3D2023} and HandOccNet \cite{parkHandOccNetOcclusionRobust3D2022}. Our study shows that self-occlusion occurs with high frequency (up to 20\%) in egocentric video of natural hand interactions and significantly negatively impacts the performance of existing SOTA hand pose estimation models, a phenomenon also identified in previous hand pose estimation systems like HandOccNet \cite{parkHandOccNetOcclusionRobust3D2022} and CHORD \cite{liCHORDCategorylevelHandheld2023}. This analysis underscores that addressing hand tracking during occlusion is not a marginal improvement, but a necessary step toward building robust hand tracking systems that enable the future of interactions that do not rely on keyboards or controllers. 

We collected a new dataset of over 170,000 high-resolution frames of dorsal hand data across 17 gestures from 12 participants. Our proposed system reduces the \changed{mean per-joint angle error (MPJAE) to SOTA models without the need for any fingers in view}, mitigates the negative impacts of self-occlusion, and is not meaningfully affected by skin color. To demonstrate the practical utility of our approach, we evaluate its performance on downstream applications like pinch and tap detection. Finally, to illustrate the potential of our skin deformation analysis, we showcase an interaction not possible with conventional egocentric methods: isometric ``force click'' detection with no discernible hand motion, akin to a trackpad press on surface or pressing fingers together from an already-touching pose.

% 5. Three main contributions - empirical analysis, technical pipeline, applications
We present several contributions:
\begin{enumerate}
    \item An analysis of the prevalence and impact of self-occlusion scenarios in common egocentric hand datasets, motivating the use of dorsal features.
    \item \systemname, an open-source end-to-end pipeline that transforms dorsal skin imagery into hand pose predictions and click detection without temporal dependencies.
    \item A 12-participant evaluation of \systemname's performance versus state-of-the-art baselines, as well as analyses with respect to occlusion, skin tone, image size, and backbone.
    \item Several exemplary use cases of \systemname in key hand interactions, including detection of pinching, tapping, and isometric force click.
\end{enumerate}

\section{Related Work}
% \subsection{Hand Tracking during Self-Occlusion}
% % how SOTA address the occlusion issues in hand tracking
% % 
Hand pose estimation has been widely studied in computer vision and HCI using either external camera setups or on-body/wearable sensors. External, environment-mounted cameras (RGB/RGB-D or multi-view) generally suffer less from self-occlusion but require infrastructure \cite{Zimmermann2017Learning, Simon2017Hand, Zhang2020MediaPipe}. In contrast, wearable cameras (head-, chest-, or wrist-mounted) provide mobility at the cost of frequent self-occlusion in egocentric views \cite{ GarciaHernando2018FirstPerson,rogez2015first}. To improve robustness to finger self-occlusion common in wearable viewpoints, we leverage dorsal-hand skin cues (tendon/wrinkle deformation) to infer occluded finger motions. Accordingly, we review methods that explicitly exploit skin appearance or deformation signals \cite{wuBackHandPose3DHand2020, yeoOpisthenarHandPoses2019}. For more comprehensive reviews on external capture systems, please refer to these surveys \cite{Erol2007review, Li2019survey, Ahmad2019review, Chatzis2020study}.

\subsection{3D Hand Pose Estimation in Egocentric and Occluded Scenarios}

Egocentric cameras (head or chest-mounted) provide a general sensing setup used not only for hand tracking but also for full-body tracking and environment understanding, so hand tracking from these views integrates naturally with broader egocentric pipelines \cite{Grauman2024Ego4D}. Often, researchers and engineers use fast prediction pipelines like MediaPipe for simple gesture recognition \cite{Zhang2020MediaPipe}. State-of-the-art (SOTA) models like HaMeR \cite{pavlakosReconstructingHands3D2023} and FrankMoCap \cite{rongFrankMocapFastMonocular2020} are trained primarily using third person data but still demonstrate strong egocentric performance. Compared to third-person views, however, egocentric perspectives are especially affected by heavy occlusion, viewpoint bias, camera distortion, and motion  blur from head movements \cite{fanBenchmarksChallengesPose2024}. To address these challenges, computer vision researchers have assembled large datasets and benchmarks tailored to egocentric hand pose estimation \cite{Kwon2021H2O, Ohkawa2023AssemblyHands, graumanEgoExo4DUnderstandingSkilled2024, fanARCTICDatasetDexterous2023, GarciaHernando2018FirstPerson}. Existing approaches attempt to mitigate some of these problems through multi-view camera capture \cite{hanMEgATrackMonochromeEgocentric2020, liuSingletoDualViewAdaptationEgocentric2024} and distortion-aware models \cite{prakash3DHandPose2024a, fanBenchmarksChallengesPose2024}, however occlusion still remains a key challenge \cite{liCHORDCategorylevelHandheld2023}.

Several models explicitly seek to address occlusion in hand tracking. HandOccNet exploits the information in occluded regions as a secondary means to enhance image features and improve hand pose estimation \cite{parkHandOccNetOcclusionRobust3D2022}. Mueller et al. tackle egocentric hand tracking under occlusion by coupling a two-stage convolutional neural network with depth-aware cropping and normalization \cite{Mueller2017RealTime}. Closer to our work, recent efforts in computer vision have focused specifically on occlusion in egocentric hand pose estimation by building large-scale datasets \cite{Ohkawa2023AssemblyHands, Kwon2021H2O, fanARCTICDatasetDexterous2023} and models \cite{Ohkawa2023AssemblyHands, Kwon2021H2O, fanARCTICDatasetDexterous2023, tseCollaborativeLearningHand2022, eldentseSpectralGraphormerSpectral2023, wangUniHOPEUnifiedApproach2025} for users interacting with occluding articulated objects,  and trajectory forecasting to predict hand pose when entering and exiting the frame of view \cite{hatanoInvisibleEgoHand3D2025}.

Like prior egocentric approaches, our method captures images from an egocentric camera and thus encounters strong self-occlusion. We directly address this problem by explicitly utilizing dorsal skin features, which encode finger position information \textit{without} the need to visually see the finger. To our knowledge, our system is the first egocentric hand pose estimation model using this approach.

\subsection{Hand Pose Estimation Leveraging Skin Features}
\label{rw_skin}
The back of the hand is particularly informative because tendons, muscles, and bones induce visible skin features closely linked to finger motion \cite{Nasir2015skin}. Biomechanical evidence further shows regional differences in dorsal skin mechanics \cite{zhaiInfluenceGraspingPostures2023}. Based on this observation, Sugiura et al. developed a wearable array of photo-reflective sensors that measures dorsal skin deformation to classify hand gestures \cite{Sugiura2017behind}. Zhao et al. developed a wearable using friction electric technology to sense dorsal deformations and predict hand gestures \cite{zhaoMachineLearningassistedWearable2024}.

Vision-based approaches observing skin features free the dorsum area from on-hand sensors. Opisthenar uses a wrist-worn camera to capture dorsal features for gesture and tap recognition \cite{yeoOpisthenarHandPoses2019}. DorsalNet uses the same camera platform with a hand simulator to capture motion images of the dorsum to predict 3D hand poses \cite{wuBackHandPose3DHand2020}. These camera-based methods all require hardware setups on the wrist or arm, making them less usable compared to the more traditional egocentric hand tracking systems. 

While prior work shows that skin features can reveal occluded hand pose, there are few systems that sense these features without close or direct contact with the dorsal area itself. Our method instead uses an egocentric (head-mounted) camera for bare-hand tracking, enabling more user freedom and better cross-applicability to existing XR devices.

\section{Motivating Study: Presence of Self-Occlusion in Egocentric Hand Tracking}
\label{sec:motivation}
We begin our work by analyzing prior egocentric datasets to understand the impact of self-occlusion in egocentric hand tracking. We aim to quantify the conditions under which dorsal features provide a viable alternative to hand sensing in the absence of traditional signals such as hand silhouettes by determining the prevalence and significance of self-occlusion in egocentric POVs.

\begin{figure}[]
    \centering
    \includegraphics[width=1\linewidth]{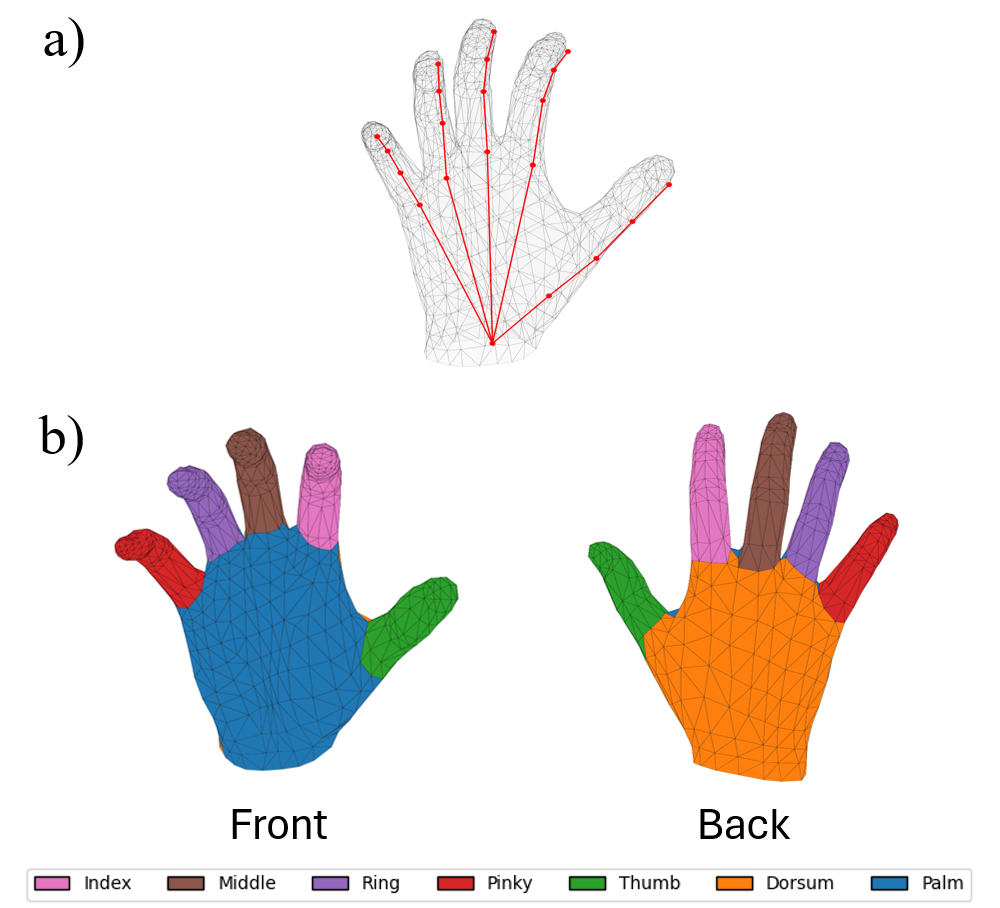}
  % Second subfigure
  \caption{a) Manually defined OpenPose keypoint alignment on the MANO model \cite{caoOpenPoseRealtimeMultiPerson2019a}. Minor deviations from the mesh are caused by MANO's internal definition of a joint position. b) Categorization of MANO mesh faces. Each color represents a separate categorization corresponding to one of index, middle, ring, pinky, thumb, dorsum, and palm.}
  \label{fig:mano-labelings}
  \Description{
  a) Image of a hand mesh with keypoints labeled. Twenty-one keypoint are connected starting from a root joint position. b) Image of a hand mesh with faces grouped by color. Groups are as follows: index finger, middle finger, ring finger, pinky, thumb, back of the hand, palm of the hand.
  }
\end{figure}

\label{sec:self-occ-measurement}
\subsection{Measuring Self-Occlusion} 

To measure the amount of hand self-occlusion in an image, we fit each pose annotation to a hand mesh using MANO \cite{MANO:SIGGRAPHASIA:2017}, a realistic non-rigid human hand model, to render the hand mesh. Using MANO, we can express any hand mesh $\Theta$ as a function of the hand pose $\theta$, hand shape parameters $\beta$, and translation $t$: $\Theta = \mathrm{MANO}(\theta, \beta) +t$. For datasets that only provide 3D OpenPose keypoints $\hat{J}$, we match the corresponding keypoints to the generated MANO joints and mesh vertices $J$. A visualization of our keypoint alignment is shown in \Cref{fig:mano-labelings}a. We then fit a hand mesh using MANO with an objective function defined as:

\begin{equation}
    \mathcal{L}_{\textrm{joints}}(\theta, \beta, t) = \big\lVert J - \hat{J}\big\rVert^2_2 + \big\lVert\theta \big\rVert^2_2 + \big\lVert \beta \big\rVert^2_2
\end{equation}

While this fitting process is not strictly optimal since we did not optimize against the hand silhouette to compute shape parameters, it still provides a sufficient understanding of the general self-occlusion of the hand in each frame. We compute the percent occlusion of each hand mesh through the following two-step process. 

Using the hand mesh $\Theta$, we project vertices $V$ into the egocentric camera space using the camera extrinsic parameters $V_{\textrm{cam}} = RV_{\textrm{world}} + t$ collected from initial camera calibration. For each face of the MANO model, we compute its normal vector and filter to only faces with a normal vector facing the camera ($-z$). We then apply Z-buffering \cite{catmullHiddensurfaceAlgorithmAntialiasing1978} using this filtered list of faces. The 3D mesh is projected into the camera's image plane where each mesh face becomes a 2D triangle. We then create an array of depth values, known as a Z-buffer, to track the closest distance to some face at that pixel. For each pixel, we compute the triangle membership through barycentric coordinate sampling and track the interpolated depth at that point. If the interpolated depth is equal or less than the stored Z-buffer depth, the face is marked as visible; otherwise it is occluded and ignored. We then compute the surface area of each visible face as a measurement of mesh visibility. This measurement only considers visibility in respect to other faces of the hand and the camera view. Thus, this only measures the amount of self-occlusion per frame and does not account for occlusion from manipulating other objects or body parts, which disproportionately affects the fingers and palm. 

\begin{table*}[]
\caption{Statistics for the four analyzed egocentric hand pose datasets. ``\# Frames Analyzed'' only counts the number of frames with a right hand within view of the egocentric camera.  ``\# Frames Occluded'' denotes the number of frames with at least one finger's surface over 90\% occluded. }
\label{tab:motivation-dataset-info}
\resizebox{.9\linewidth}{!}{%
\begin{tabular}{@{}lcccccc@{}}
\toprule
Dataset & \# of Subjects & Annotation & Gesture Type & In the Wild & \# Frames Analyzed & \# Frames Occluded \\ \midrule
ARCTIC \cite{fanARCTICDatasetDexterous2023} & 10 & MANO & Articulating Objects & \ding{55} & 184,346 & 25,242 \\
H2O \cite{Kwon2021H2O} & 4 & MANO & Articulating Objects & \ding{55} & 26,529 & 5,666 \\
EgoExo4D \cite{graumanEgoExo4DUnderstandingSkilled2024} & 740 & Keypoints & Skilled Activities & \ding{51} & 8,703 & 1,873 \\
AssemblyHands \cite{Ohkawa2023AssemblyHands} & 34 & Keypoints & Assembling Toys & \ding{55} & 394,621 & 140,669 \\ \bottomrule
\end{tabular}
}
\end{table*}

\subsection{Occlusion Prevalence in Egocentric Data}
Using our occlusion metric, we compute the self occlusion of the fingers for any frame with a fully visible right hand from four egocentric hand pose datasets: ARCTIC \cite{fanARCTICDatasetDexterous2023}, H2O \cite{Kwon2021H2O}, EgoExo-4D \cite{graumanEgoExo4DUnderstandingSkilled2024}, and AssemblyHands \cite{Ohkawa2023AssemblyHands}. More information on these datasets can be found in \Cref{tab:motivation-dataset-info}. Occluded faces are grouped and categorized according to seven different categories: index, middle, ring, pinky, thumb, palm, dorsum to measure the occlusion of different parts of the hand. A visualization of our categorization can be seen in \Cref{fig:mano-labelings}b. We note that for the fingers, fully visible is defined as $50\%$ of faces visible and scaled accordingly since we do not separate the top and bottom faces like for the dorsum and palm. We define fully occluded as 10\% of surface area visible to account for small errors in Z-buffer rasterization while ensuring that the majority of the finger beyond a small subsection of the proximal phalanx is occluded.

\begin{figure}
    \centering
    \includegraphics[width=.9\linewidth]{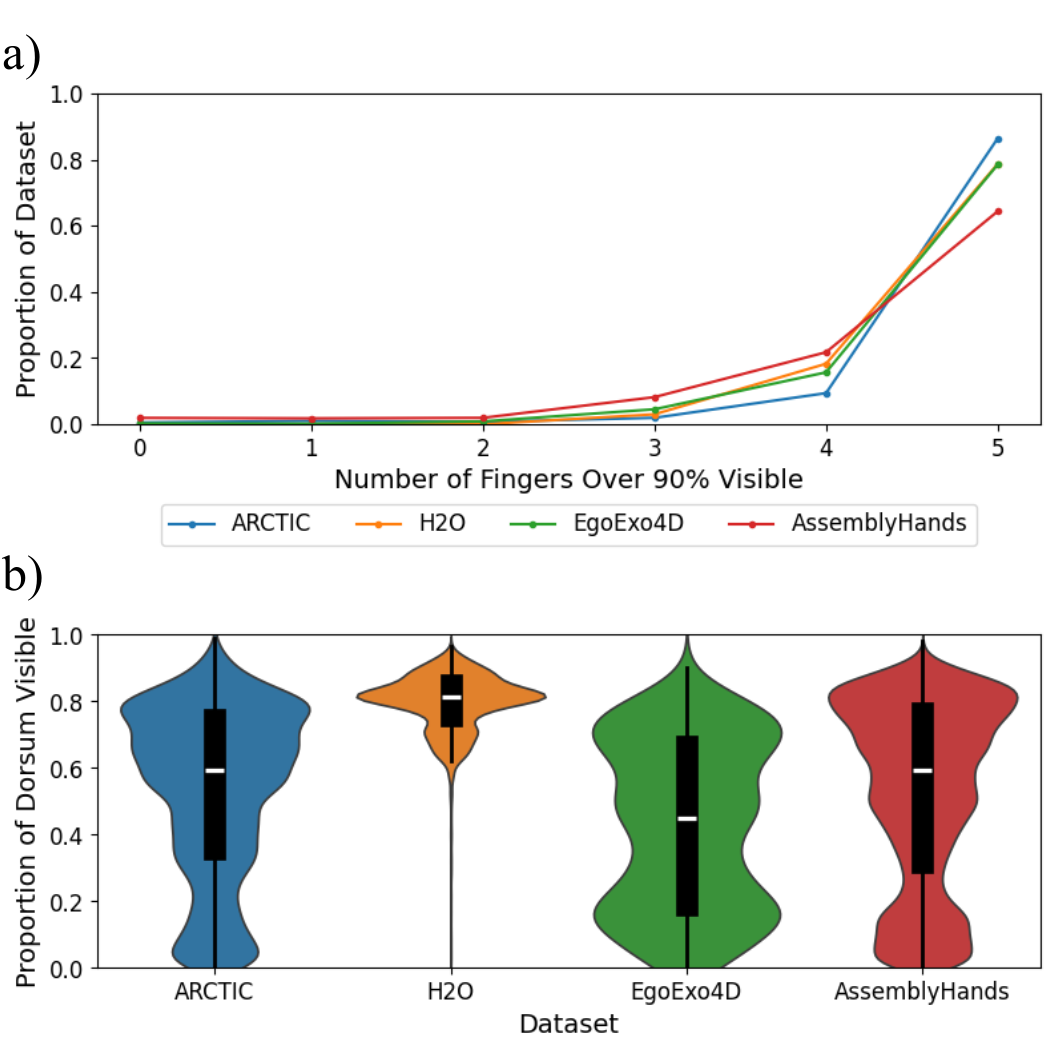}
    \caption{a) Number of fingers with over 90\% of surface faces visible across different datasets. b) Distribution of dorsal visibility when atleast one finger is over 90\% occluded across different datasets. The internal box represents the 25th to 75th percentile while the white line represents the median.}
    \label{fig:motivation-dataset-stats}
    \Description{
    a) Line graphs for four datasets showing the proprotion of the dataset with 0-5 fingers over 90\% visible. For all lines, 20\% of each dataset has at least one finger occluded. b) Violin plot showing the distribution of the dorsum of the hand visible for each dataset. Medians all fall above 40\% with the highest being 80\% (H2O) and lowest being EgoExo4D (43\%). Distributions skew towards a high percentage of visibility for all datasets.
    }
\end{figure}

\textbf{Hand self-occlusion.} 
As shown in \Cref{fig:motivation-dataset-stats}a, more than 20\% of frames across all evaluated datasets exhibit at least one occluded finger, with over 5\% containing two or more occluded fingers. These values are a conservative estimate as we do not consider any articulating objects, which are far more likely to occlude the fingers than the dorsum of the hand. These self-occlusion cases often occur because of three primary scenarios: The hand is angled downwards such that the back of the hand occludes the fingers, the hand is angled out such that the index finger occludes the other fingers, or the hand is partially out of frame. Our analysis indicates that there is a sizable proportion of egocentric hand capture data affected by self-occlusion which follows findings from literature. \cite{fanBenchmarksChallengesPose2024}.

\textbf{Hand dorsum visibility.} 
We then analyze how much of the dorsum is visible in these situations by computing the distribution of dorsal face visibility with at least one finger over 90\% occluded. We find that for every dataset, the median dorsal visibility is above 40\% and up to 80\%, indicating that a significant proportion of frames with occluded fingers still have visible dorsal features. Thus, dorsal features offer an alternative rich and stable signal when hand and finger silhouettes are not available. 

\subsection{Impact of Occlusion on Egocentric Hand Pose Estimators}
\begin{figure}
    \centering
    \includegraphics[width=.9\linewidth]{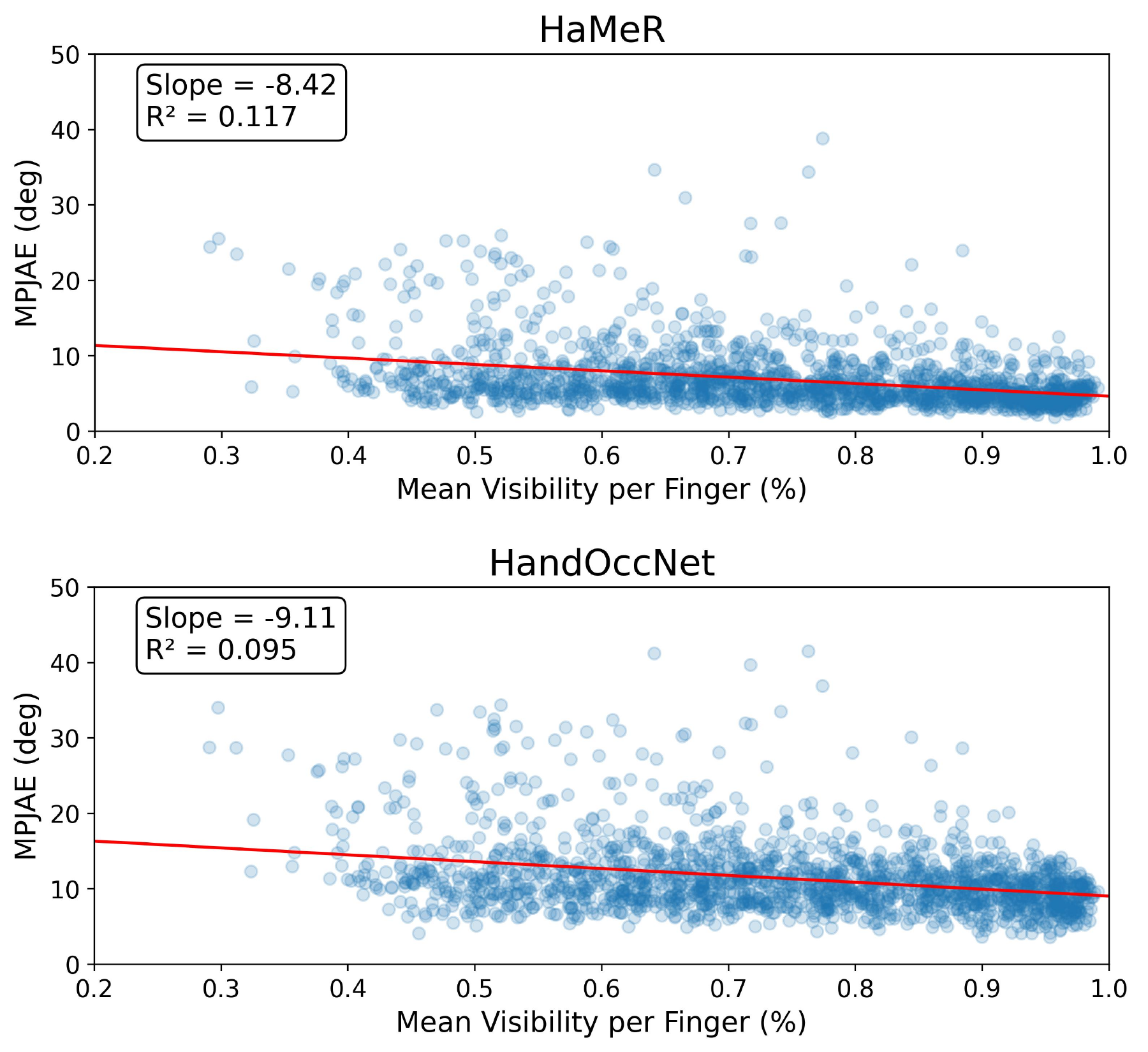}
    \caption{\changed{MPJAE compared to average per finger visibility in our unseen dataset. The red line denotes the fitted linear regression for visualized data. Mean finger visibility is defined as the percentage of face surface area that is visible to the camera averaged across all five fingers. The higher the MPJAE, the worse the performance of the model is.}}
    \label{fig:baseline-perf-occ-motivation}
    \Description{
    Two scatter plots of mean visibility per finger against MPJAE in degrees for HaMeR and HandOccNet. Regression lines for each plot show the decrease in performance with occlusion. HaMeR showed the steepest with slope=-15.84 and R^2=0.212 followed by HandOccNet with slope=-9.11 and R^2=0.95.
    }
\end{figure}

Following our findings showing that self-occlusion is often present in egocentric views, we analyzed the impact that self-occlusion has on the performance of existing hand pose estimation models. We collected a ground truth dataset of 17 gestures from 12 participants for 170k frames of egocentric hand pose data with hand meshes generated from an accurate motion capture system and evaluate two baseline models on this dataset. More information on our dataset can be found in \Cref{sec:dataset}. \changed{We evaluate the performance of the best performing checkpoint of two state-of-the-art baselines, HaMeR \cite{pavlakosReconstructingHands3D2023}, a hand mesh estimator using a fully transformer-based architecture, and HandOccNet \cite{parkHandOccNetOcclusionRobust3D2022}, an occlusion-robust hand mesh estimator exploiting information in occluded regions, on a completely unseen dataset described in \Cref{sec:dataset}.} We then quantify pose prediction performance using mean per joint angular error (MPJAE) against our motion capture ground truth and plot this error against the mean per finger visibility computed through our occlusion metric described in \Cref{sec:self-occ-measurement}. Our results are shown in \Cref{fig:baseline-perf-occ-motivation}.

We find that for both HaMeR and HandOccNet, there is a statistically significant ($p<.0001$) negative correlation between finger visibility and error ($m=-15.84\degree$, $R^2=0.212$ for HaMeR, $m=-9.11\degree$, $R^2=0.095$ for HandOccNet). This is especially apparent in HaMeR, which has been widely regarded as a standard in pose estimation since its release and has become the backbone of many dataset annotation and hand pose estimation systems \cite{zhaoEgoPressureDatasetHand2024a, prakash3DHandPose2024a}. Notably for HandOccNet, we found that the model would often default to outputting a neutral hand position in self-occlusion situations, leading to serious prediction errors most likely due to the model's limited precision in predicting hands occluded behind other objects. Our findings are consistent with our intuition that most existing hand tracking models rely on the general silhouette that lacks the information fidelity to predict pose. Often, their architecture prevents them from using skin features due to the limited token and image size allowed.

\subsection{Our Proposal: Investigate Dorsal Features}
Given our findings, we choose to investigate dorsal features as a key signal to improve hand pose estimation in egocentric POVs. We lay out our work as follows: we first perform a high-resolution data collection of dorsal features from different hand gestures to capture and analyze dorsal features in detail. Using this dataset, we propose and train a new model that normalizes visual features from the dorsum to a neutral hand pose, similar to how DorsalNet uses motion images \cite{wuBackHandPose3DHand2020}, to isolate dorsal features even through global motion. We then evaluate our system against  baseline hand pose estimation models and across skin tones and image resolutions. Finally, we demonstrate the utility of dorsal features beyond pose estimation by using them in the detection of interaction modalities including pinch, tap, and click.

\begin{figure*}
    \centering
    \includegraphics[width=.98\linewidth]{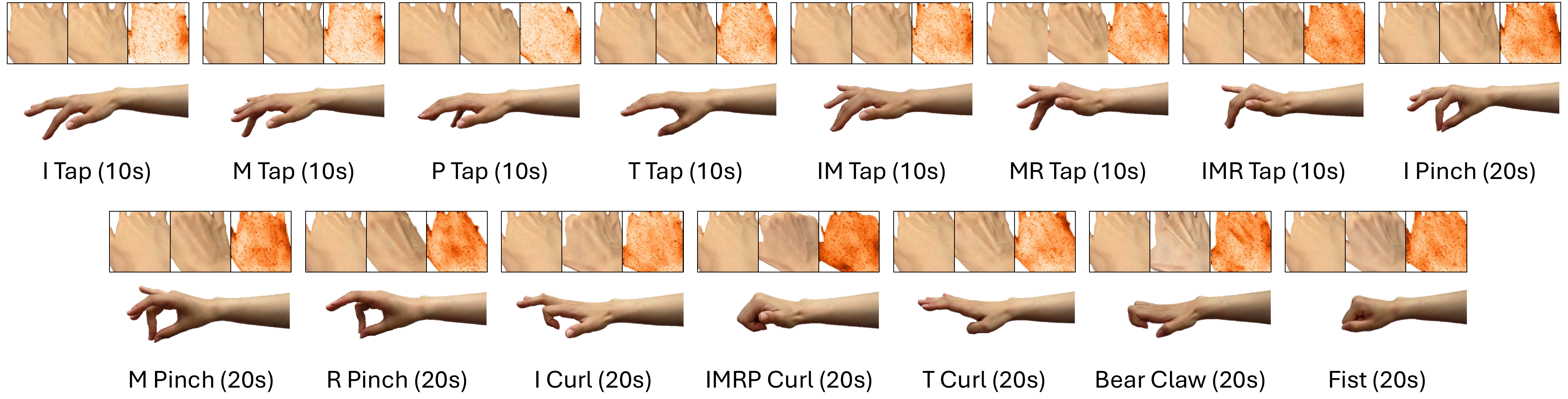}
    \caption{Examples of each static gesture collected in our data collection. \changed{Not depicted are the two dynamic gestures: fanning (30s) and freeform (60s)}. On the top are the following: An aligned image of the reference, the picture of the dorsal features during this gesture, and the cosine similarity mapping for the DINO features generated from the reference and the current image. The color of the similarity map indicates a smaller cosine similarity (darker is more different). (I: index finger, M: middle finger, R: ring finger, P: pinky, T: thumb).}
    \label{fig:gesture-info}
    \Description{
    Visualization of 15 gestures with an image of the back of the hand in a neutral pose, during the gesture, and a cosine similarity map comparing the two images. Gestures are as follows: I Tap (10s), M Tap (10s), P Tap (10s), T Tap (10s), IM Tap (10s), MR Tap (10s), IMR Tap (10s), I Pinch (20s), M Pinch (20s), R Pinch (20s), I Curl (20s), IMRP Curl (20s), T Curl (20s), Bear Claw (20s), Fist (20s). For smaller gestures like taps, we find small changes in the similarity map primarily centering around the knuckles and tendons of the hand. For pinches, we see large differents appear towards the bottom of the dorsal area. For curls, bear claw, and fist, we see large differences appear across the entire dorsal area. 
    }
\end{figure*}

\section{High-Resolution Dorsal Data Collection}
\label{sec:dataset}
Since dorsal and skin deformation are often very small and require a higher resolution than traditional hand pose datasets, we collected our own dataset of 4K images of dorsal features from an egocentric point of view (POV) to mimic natural hand self-occlusion. To reduce prediction inaccuracies and enable the capture of fine-grained gesture motions, we used a Vicon Motion Capture setup with 8 Vero cameras \cite{ViconVeroAdvanced} and a custom marker set shown in \Cref{fig:vicon-marker-def} to annotate 3D hand poses. Motion capture data was collected in tandem with the image feed from an iPhone 12 Pro Max recording at $3840\times2160$ and 30 FPS to mimic the image quality of a mobile, consumer camera. Camera intrinsics and extrinsics were calibrated through a ChArUco board with markers to identify board corners in motion capture space. The camera was stationary to control for motion blur given our low capture rate and to best isolate dorsal textures. While cameras were stationary, users were allowed to move their hand around the camera frame of view. Examples of the high-resolution dorsal images are shown in \Cref{fig:gesture-info}.

\subsection{Participants and Procedure}
We recruited 12 participants (6M, 6F) between the ages of 18-29 with Monk Skin Tone Levels \cite{SkinToneResearch} ranging from 2-7. More information on our participants can be found in \Cref{tab:participant-info}. We first collected a 10-second capture of their right hands' dorsal area while in a neutral and relaxed position pose for calibration. Participants were then asked to perform 17 different dynamic gestures and freeform hand poses. \changed{Gestures were selected by surveying common XR interactions and gestures that commonly exhibit high levels of self-occlusion in prior data. We acknowledge that this is a limited subset of all possible hand poses.} For each gesture, participants started from a relaxed position and performed the specific gesture repeatedly (freeform hand poses had no limitations). We asked users to perform gestures at a normal and controlled speed for all captures. More information on our collected gestures is shown in \Cref{fig:gesture-info}. During the study, participants placed their right arm on an armrest with their hand approximately 20 cm away from the camera with the camera at a 45 degree angle from the hand to mimic an egocentric camera's position. Participants were asked to repeat these gestures for three different hand positions to mimic different dorsal and finger occlusion scenarios: hand pointed straight, hand pointed downwards at a 30 degree angle, and hand turned out at a 30 degree angle for a total of 51 separate trials. \changed{Our dataset has comparable diversity of participants and size to previous egocentric hand pose datasets including HO3D \cite{hampaliHOnnotateMethod3D2020a}, FreiHAND \cite{Freihand2019}, EgoHands \cite{Bambach_2015_ICCV}, and ARCTIC \cite{fanARCTICDatasetDexterous2023} while being the first 4K egocentric dataset.} All participants were compensated with $\$25$ for their time. Our study was approved by our institution's IRB.

\subsection{Data Cleaning}
During data collection, we found that the motion capture system could mislabel markers in the solving process. Thus, researchers manually evaluated all data, fixed any marker swapping, and removed any poorly predicted frames due to marker occlusion or system error. Researchers then manually aligned motion capture and video streams using a clapper board and transformed all markers from motion capture space into camera space using the calibration extrinsics and intrinsics. We then computed the MANO parameters for each trial.

Since we only evaluate hand poses and not the full mesh, we obtain an initial estimate of the shape parameters for each participants using HaMeR \cite{pavlakosReconstructingHands3D2023} with each participant's neutral hand capture, which shows the entire hand, to mimic a realistic implementation scenario. Using these shape parameters, we associate motion capture marker positions $\hat M$ with a corresponding MANO mesh vertex $M$. We then fit the MANO model using the following objective function:

\begin{equation}
    \mathcal{L}_{\textrm{markers}}(\theta, t) = \big\lVert M - \hat{M}\big\rVert^2_2 + \big\lVert\theta \big\rVert^2_2 
\end{equation}

In total, we collected 172,222 frames of annotated pose data for 17 gestures across 12 participants.
\begin{figure*}
    \centering
    \includegraphics[width=.99\linewidth]{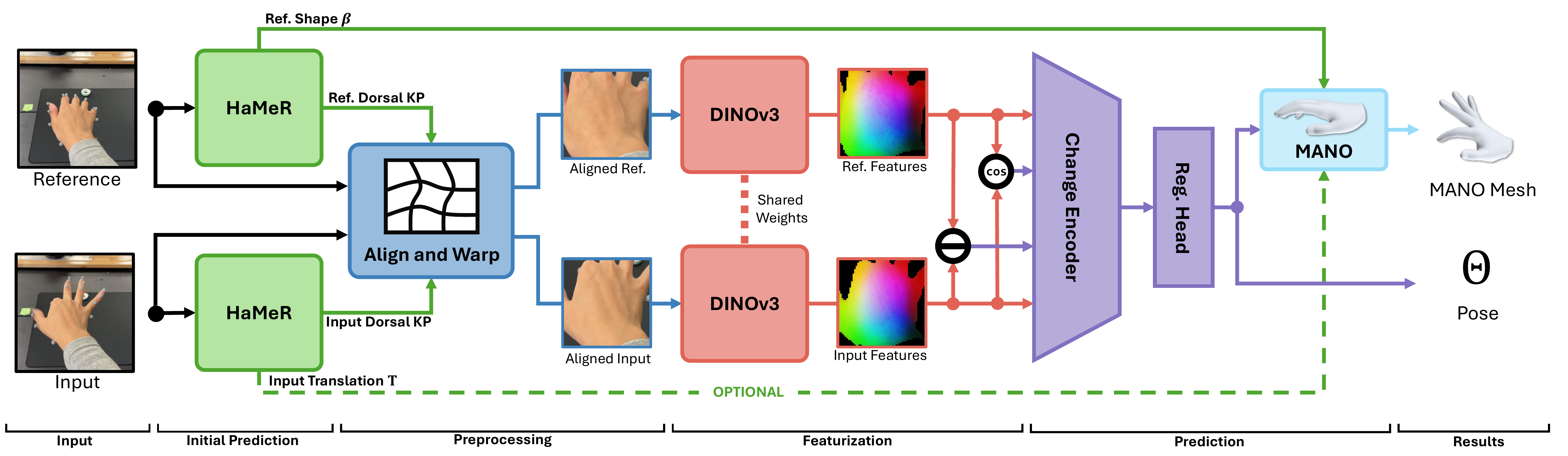}
    \caption{\systemname's system architecture. Users input a ``reference'' image of their hand in a neutral position and a picture of the hand in some gesture. An initial hand pose prediction from HaMeR is then used to align both hands so that their dorsal features are spatially localized. Images of dorsal features are then fed into DINOv3 to extract image features. These features, along with the cosine similarity and difference between the ``reference'' and current image's features, are fed into the change encoder. A regression head then predicts the current hand pose, which can be processed with MANO using a prior shape prediction to generate a hand mesh. Optionally, users can use the initial translation prediction from HaMeR to localize the final mesh in the camera frame.}
    \label{fig:system-architecture}
    \Description{
    System diagram of DeltaDorsal. A reference and input image are inputted to HaMeR. Predictions from HaMeR are used to align and warp dorsal features which are separate fed into DINOv3 featurizers with the same weights. Features are outputted which are subtracted with cosine similarity and inputted into a change encoder and regression head. The head outputs pose which can be used to create a MANO mesh with the reference shape and, optionally, the input translation. 
    }
\end{figure*}

\section{Design and Implementation of \systemname}

In this section, we detail the system design and machine learning implementation of \systemname: our pipeline to extract and leverage visual features from the dorsum of a hand to predict a full hand pose from a monocular egocentric view. Notably, we propose a time-invariant model that only requires a single initialization process for each user.

To represent the hand, we use MANO \cite{MANO:SIGGRAPHASIA:2017}. Since the dorsum of the hand itself does not tell us any information about the localization or shape of the hand, we set the global orientation, $t$, to zero and compute the shape, $\beta$ from an initial capture of the hand. Thus, our machine learning model only predicts the 15 joint angles that comprise the pose parameter, $\theta$, which can be used to resolve a mesh, $\Theta$, with a precomputed shape and translation.

\subsection{Data Preprocessing}

Following Wu et al. \cite{wuBackHandPose3DHand2020}, we found it difficult to observe subtle dorsal deformations using a singular frame. Rather than using motion images, which rely on temporal features which may disconnect or misalign with use, we compare every user's dorsal images to an image of the dorsum of their hand at a neutral position called a \textit{``reference''}. Using HaMeR \cite{pavlakosReconstructingHands3D2023}, we predict the 3D joint positions of the back of the hand and project them into the 2D camera plane using the camera intrinsics and predicted extrinsics. We then align the reference to the captured hand image by a homography estimation using RANSAC \cite{fischlerRandomSampleConsensus1981} on the projected 2D points. Images are cropped to just the dorsum of the hand and resized to $384\times384$ for training. This procedure localizes both the reference and captured image so that both hands occupy the same spatial location for direct pixel-to-pixel comparison. Furthermore, by using just the dorsal features, we limit the amount of inputted data to our machine learning model to lower computational complexity and focus training on dorsal features over the hand silhouette. An example of this data is shown in \Cref{fig:system-architecture}. We note that while we use HaMeR here for simplicity and congruency with later analyses, any 2D or 3D hand pose estimator or segmentor could be used to a similar effect. For training purposes, we add random brightness, contrast, gamma, noise, and gaussian blur to better generalize to different lighting and motion conditions. \changed{Augmentations are performed independently for the input and reference images to improve generalizability across different lighting and hand conditions.}

\subsection{Network Architecture}

\systemname adopts a two-stream architecture built on a transformer backbone with a lightweight convolutional change encoder followed by a regression head. A visualization of this architecture is shown in \Cref{fig:system-architecture}. Using a pretrained DINOv3 Vision Transformer (ViT) \cite{simeoniDINOv32025} following the ``large'' design (ViT-L) with the last three blocks unfrozen, we first compute the dense image features for both the reference $I_0$ and target $I_t$ using the image of the dorsum of the hand, $F_0$ and $F_t$ respectively with dimensions equal to the original image size divided by the patch size ($16\times 16$).  Notably, model weights are shared across each stream to ensure that $F_0$ and $F_t$ maintain the same feature space. A change encoder then processes this feature map by fusing the feature delta, $F_t-F_0$, cosine similarity per patch, $\cos (F_t, F_0)(x,y)$, $F_t$, and $F_0$ and feeding this data through a small convolutional network to generate the encoded features $X$. Cosine similarity features are visualized in \Cref{fig:gesture-info}. By fusing the original features, feature deltas, and cosine similarities in the change encoder, we allow the model to select between general image features and changes in dorsal features as needed. We settled on a dense featurizer backbone after preliminary experiments with optical flow methods like RAFT \cite{teedRAFTRecurrentAllPairs2020} and Gunnar-Farneback optical flow \cite{farnebackTwoFrameMotionEstimation2003} had difficulty separating skin deformations from global movements of the hand. Furthermore, traditional feature extractors like ResNet \cite{heDeepResidualLearning2015} and ConvNeXt \cite{liuConvNet2020s2022} failed to generalize to out-of-distribution subjects. We believe this limitation arises because these models are not well-suited to capture the subtle soft-body surface deformations in the back of the hand which often do not demonstrate distinct, localized feature changes. Finally, a regression head applies a small convolutional neck for local refinement, pools spatial features, and passes pooled vectors through a shallow multi-layer perceptron to predict the pose vector $\beta$ ($(15,3)$ for 15 axis-angle rotations). 

Using our predicted pose parameter $\theta$ and the shape parameter computed by HaMeR on the reference image $\beta_0$, we compute the MANO mesh and predicted 3D joints $J$ shown in \Cref{fig:mano-labelings}a. Following the procedures defined in Pavlakos et al. \cite{pavlakosReconstructingHands3D2023}, we directly apply a loss on the pose parameter $\beta$ and encourage consistency in the 3D space by supervising the 3D joint positions $\hat J$. Our full loss can be formalized as

\begin{equation}
    \mathcal{L}_{\beta} = \big\lVert J - \hat{J}\big\rVert_1 + \big\lVert \theta - \hat{\theta}\big\rVert^2_2
\end{equation}

\subsection{Implementation Details}
All systems were implemented using PyTorch \cite{paszkePyTorchImperativeStyle2019}. We used the AdamW optimizer \cite{loshchilovDecoupledWeightDecay2019} with a batch size of 96 for our training. During training, we randomly selected a reference frame from the corresponding calibration data capture. Please refer to our codebase and configuration files for specific parameter settings for experiments. All training and evaluations were performed using PyTorch Distributed Data Parallel on a system with an AMD Ryzen Threadripper PRO 3955WX 3.90GHz, four Nvidia RTX A5500s, and 256GB of RAM. Our full codebase can be found at \url{https://github.com/hilab-open-source/deltadorsal}.

\label{sec:evaluation}
\section{Evaluation}
We evaluate \systemname's 3D hand pose estimation performance against SOTA vision-based hand pose estimation models. The following documents our procedure and results.

\begin{figure*}
    \centering
    \includegraphics[width=0.98\linewidth]{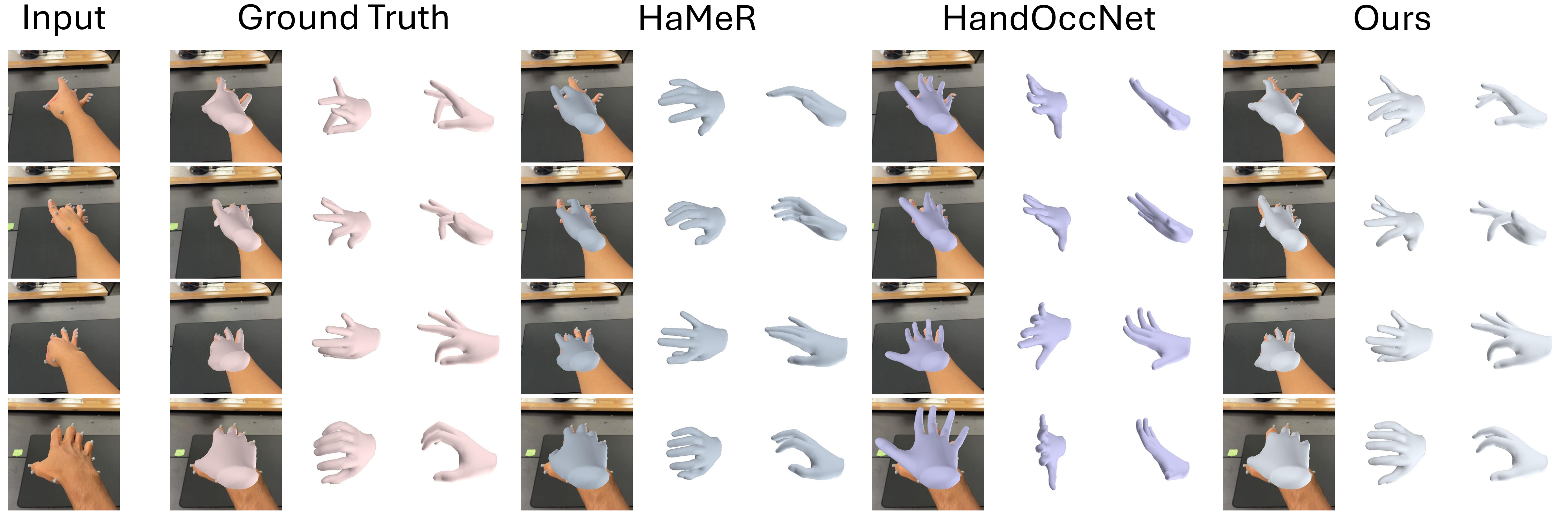}
    \caption{\changed{Examples of ground truth and predicted hand mesh from HaMeR, HandOccNet, and our system in self-occluded scenarios. For both baselines, we inputted a full image of the hand while our system only used the cropped dorsal features of the hand.}}
    \label{fig:prediction-examples}
    \Description{
    Visualization of the predictions from HaMeR, HandOccNet, and DeltaDorsal. HandOccNet outputs a neutral flat hand for all input images. For images where the fingers are occluded, HaMeR outputs a flatter hand with open fingers where DeltaDorsal accurately predicts the other fingers. When fingers are visible, HaMeR accurate predicts their pose while DeltaDorsal performs similarly. 
    }
\end{figure*}

\label{sec:perf-study}
\subsection{Pose Estimation Performance Study}
We first conduct a baseline performance study to evaluate our model and compare performance with existing computer vision methods using the full hand silhouette. For all experiments, we adopt a leave-one-subject-out (LOSO) cross-validation paradigm ($\approx$12,000 test frames per subject), where each participant was held out in turn for testing while remaining participants were used for training, to assess cross-subject generalization and isolate performance gains attributable to dorsal movement representation rather than subject-specific hand features. We adopt the mean per joint angular error (MPJAE) metric to evaluate the performance of our pose prediction. We follow the evaluation protocols of prior 3D hand pose works \cite{hampaliHOnnotateMethod3D2020a, pavlakosReconstructingHands3D2023} and report the Procrustes-aligned mean per-joint position error (PA-MPJPE) of our system using the initial shape parameter estimated from calibration. This metric measures joint error after global translation, rotation, and scale alignment to evaluates our system's mesh reconstruction performance in realistic application scenarios. 

\changed{We compare our method with two leading hand-pose estimators, HaMeR \cite{pavlakosReconstructingHands3D2023} and HandOccNet \cite{parkHandOccNetOcclusionRobust3D2022}. HaMeR reports state-of-the-art (SotA) or near SotA results across multiple datasets and notably uses a ViT-H backbone (632M parameters), more than twice the size of our ViT-L backbone (300M). HandOccNet is explicitly designed to be occlusion-aware: it first predicts features for occluded regions and then infers pose. To ensure a fair comparison, we fine-tuned both models under the same leave-one-subject-out (LOSO) protocol, freezing the backbone and training the prediction head to convergence. Notably, this differs from our training strategy for \systemname which unfroze the last three layers of the backbone as well to maintain the strong learned featurization and generalization of the model's pretraining on larger datasets while mitigating convergence and instability issues in our smaller backbone compared to HaMeR (VIT-H). Both models were given the full image of the hand including the fingers compared to just the dorsal crop of \systemname. In our experiments, HandOccNet proved difficult to fine-tune or train end-to-end and often collapsed. We suspect this reflects an architectural bias toward specific non-self-occlusion scenarios. For completeness, we still report all metrics.}
Our quantitative metrics are shown in \Cref{tab:performance-results} and examples of predicted hand meshes are shown in \Cref{fig:prediction-examples} and \Cref{fig:teaser}. 

\begin{table*}[]
\caption{Average results of leave-one-subject-out experiments across all participants for each finger. Joints are ordered from closest to furthest away from the wrist using the OpenPose keypoint labels (\Cref{fig:mano-labelings}a). MAE indicates mean angular error expressed in degrees. ($\text{Mean}_{\text{participants}} \pm \text{SD}_\text{participants}$).}
\label{tab:performance-results}
% \resizebox{.55\linewidth}{!}{%
\small
\changed{
\begin{tabular}{l|l|ccc}
Finger & Joint Metric & HaMeR \cite{pavlakosReconstructingHands3D2023} & HandOccNet \cite{parkHandOccNetOcclusionRobust3D2022} & Ours \\ \hline
\multirow{3}{*}{Index} & MCP MAE ($\degree$) & \phantom{1}$7.38 \pm 2.07 $ & $ 10.61 \pm 3.25 $ &\phantom{1}$ \mathbf{6.81 \pm 1.42} $ \\
 & PIP MAE    ($\degree$) & \phantom{1}$ \mathbf{8.10 \pm 1.11} $ & $ 16.02 \pm 3.00 $ & \phantom{1}$8.84 \pm 2.81 $ \\
 & DIP MAE    ($\degree$) & \phantom{1}$ 8.39 \pm 3.57 $ & $ 10.90 \pm 3.29 $ & \phantom{1}$ \mathbf{6.92 \pm 4.01} $ \\ \hline
\multirow{3}{*}{Middle} & MCP MAE ($\degree$) & \phantom{1}$\mathbf{7.43 \pm 1.28} $ & $ 11.38 \pm 1.77 $ & \phantom{1}$ 7.69 \pm 1.47 $ \\
 & PIP MAE    ($\degree$) & \phantom{1}$\mathbf{7.09 \pm 1.26} $ & $ 16.08 \pm 4.46 $ & \phantom{1}$ 8.32 \pm 2.82 $ \\
 & DIP MAE    ($\degree$) & \phantom{9}$ \mathbf{4.28 \pm 0.72} $ & $ 10.71 \pm 2.16 $ & \phantom{1}$ 4.47 \pm 0.97 $ \\ \hline
\multirow{3}{*}{Ring} & MCP MAE ($\degree$) & \phantom{1}$ \mathbf{6.39 \pm 0.91} $ & \phantom{1}$ 9.35 \pm 1.81 $ & \phantom{1}$ 6.90 \pm 1.78 $ \\
 & PIP MAE    ($\degree$) & \phantom{1}$ \mathbf{6.53 \pm 1.20} $ & $ 14.27 \pm 2.76 $ & \phantom{1}$ 7.07 \pm 2.24 $ \\
 & DIP MAE    ($\degree$) & \phantom{1}$ \mathbf{4.58 \pm 1.70} $ & \phantom{1}$ 9.42 \pm 2.31 $ &\phantom{1} $ 5.16 \pm 2.11 $ \\ \hline
\multirow{3}{*}{Pinky} & MCP MAE ($\degree$) & \phantom{1}$ 8.42 \pm 1.67 $ & $ 12.00 \pm 2.24 $ & \phantom{1}$ \mathbf{7.35 \pm 1.30} $ \\
 & PIP MAE    ($\degree$) & \phantom{1}$ 6.96 \pm 1.11 $ & \phantom{1}$ 9.76 \pm 3.17 $ & \phantom{1}$ \mathbf{6.28 \pm 2.45} $ \\
 & DIP MAE    ($\degree$) & \phantom{1}$ 3.82 \pm 1.38 $ & \phantom{1}$ 6.58 \pm 1.62 $ & \phantom{1}$ \mathbf{3.62 \pm 1.30} $ \\ \hline
\multirow{3}{*}{Thumb} & CMC MAE ($\degree$) & \phantom{1}$ 5.70 \pm 0.73 $ & $ 14.29 \pm 4.34 $ & \phantom{1}$ \mathbf{5.52 \pm 1.61} $ \\
 & MCP MAE    ($\degree$) & \phantom{1}$ 7.38 \pm 1.66 $ & \phantom{1}$ 9.11 \pm 2.35 $ & \phantom{1}$ \mathbf{4.77 \pm 0.63} $ \\
 & IP MAE    ($\degree$) & \phantom{1}$ 8.63 \pm 2.64 $ & \phantom{1}$ 8.57 \pm 2.17 $ & \phantom{1}$ \mathbf{6.38 \pm 1.59} $ \\ \hline
\multirow{2}{*}{Overall} & MPJAE ($\degree$) & \phantom{1}$ 6.74 \pm 0.48 $ & $ 11.27 \pm 0.97 $ & \phantom{1}$ \mathbf{6.41 \pm 0.92} $ \\
 & PA-MPJPE (mm) & \phantom{1}$ \mathbf{7.11 \pm 0.47}$ & $ 14.38 \pm 1.22$ & \phantom{1}$ 7.41 \pm 1.01$
\end{tabular}
% }
}
\end{table*}

\changed{
\textbf{Comparing against baselines.}
When comparing against our two baselines, HaMeR and HandOccNet, we find that our approach leveraging just dorsal features of the hand outperforms both baseline models in pose accuracy (MPJAE: Ours: $6.41\pm0.92\degree$, HaMeR: $6.74\pm0.48\degree$, HandOccNet: $11.27\pm0.97\degree$) and matches hand mesh pose performance even with a suboptimal shape fit (PA-MPJPE: Ours: $7.41\pm1.01 \text{mm}$, HaMeR: $7.11\pm0.47 \text{mm}$, HandOccNet: $14.38\pm1.22 \text{mm}$) (\Cref{tab:performance-results}). We note that HandOccNet often predicts a neutral hand position during self-occluded scenarios (\Cref{fig:prediction-examples}), potentially due to high amount of self-occlusion and architectural incompability. Notably our system demonstrates stronger performance for the thumb. Since most of our collected data was from an egocentric top-down POV, the thumb was disproportionately occluded and subsequently performed worse in silhouette-based models (Thumb MCP MPJAE: Ours: $4.77\pm0.63\degree$, HaMeR: $7.38\pm1.66\degree$, HandOccNet: $9.11\pm2.35\degree$). Dorsal skin deformations are induced more evenly across all fingers, enabling our approach to maintain roughly even performance across all fingers. 
}

% \textbf{Joint and finger performance.} 

\begin{figure}
    \centering
    \includegraphics[width=0.98\linewidth]{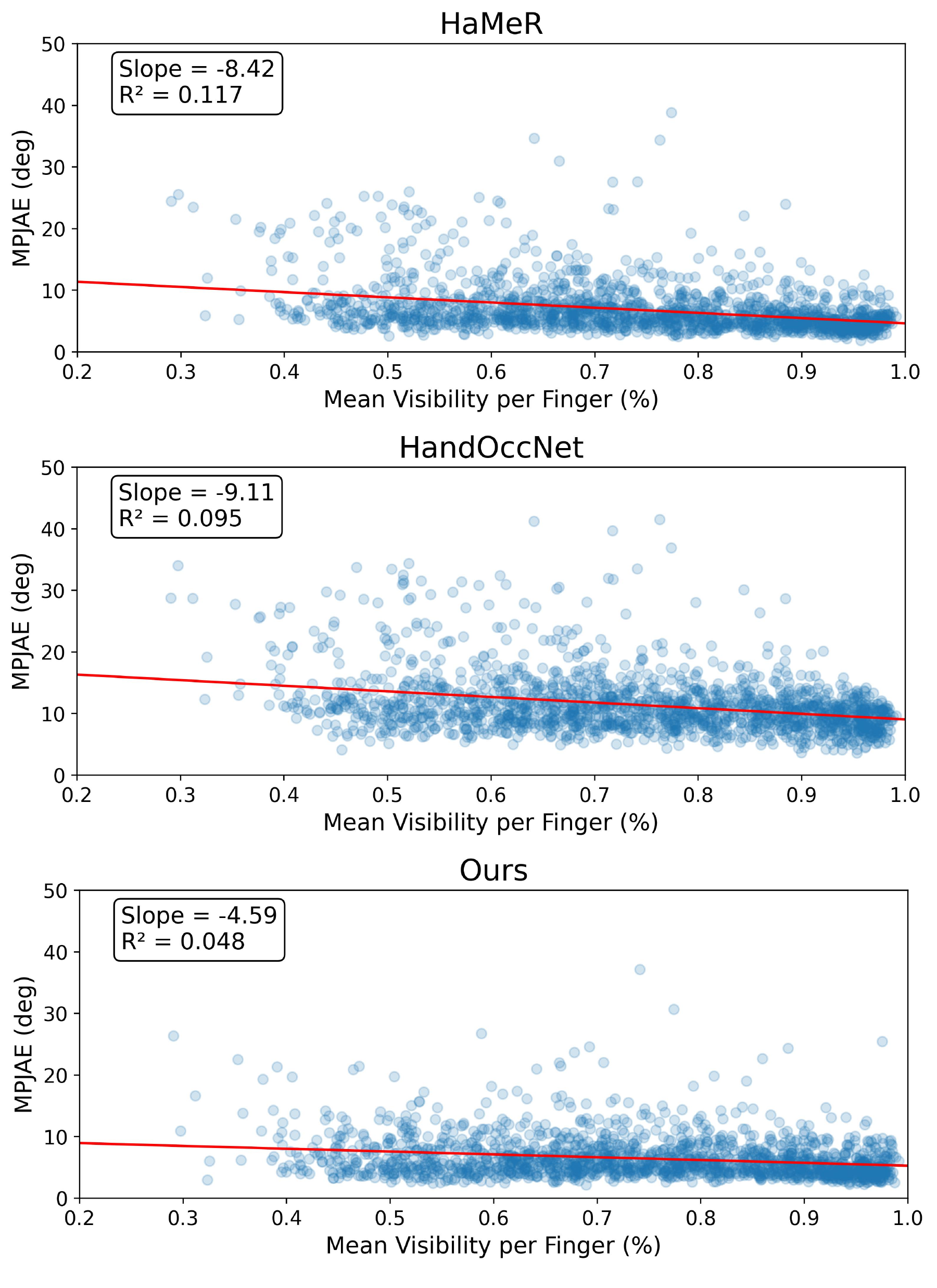}
    \caption{\changed{MPJAE compared to average per finger visibility in fine-tuned models. The red line denotes the fitted linear regression for visualized data. Mean finger visibility is defined as the percentage of face surface area that is visible to the camera averaged across all five fingers. The higher the MPJAE, the worse the performance of the model is.} }
    \label{fig:baseline-perf-occ}
    \Description{
    Three scatter plots of mean visibility per finger against MPJAE in degrees for HaMeR, HandOccNet, and DeltaDorsal. Regression lines for each plot show the decrease in performance with occlusion. HaMeR showed the steepest with slope=-8.42 and R^2=0.117 followed by HandOccNet with slope=-9.11 and R^2=0.95, and finally DeltaDorsal with slope=-.459 and R^2=0.048.
    }
\end{figure}

\changed{
\textbf{Performance across occlusion levels.}
We also evaluate the performance of our baselines and system under different occlusion levels after fine-tuning. As shown in \Cref{fig:baseline-perf-occ}, we find that our model has a smaller dropoff in performance compared to both HaMeR and HandOccNet ($m=-4.59\degree$, $R^2=0.048$ for ours, $m=-8.42\degree$, $R^2=0.117$ for HaMeR, $m=-9.11\degree$, $R^2=0.095$ for HandOccNet)\footnote{We denote slope as $m$}. HandOccNet's general performance was not significantly changed after finetuning due to previously mentioned training difficulties. When comparing performance at 50\% finger visibility and lower, \systemname outperforms HaMeR across both MPJAE (Ours: $7.59 \pm 4.38\degree$, HaMeR: $9.24 \pm 5.54\degree$, percent change: $-17.86\%$) and PA-MPJPE (Ours: $8.47 \pm 3.84$mm, HaMeR: $9.03\pm3.65$mm, percent change: $-6.20\%$). These findings are consistent with our intuition that existing hand pose estimation models rely heavily on finger silhouettes for pose prediction. Examples of this poor performance are shown in \Cref{fig:teaser} and \Cref{fig:prediction-examples}, where HaMeR and HandOccNet both fail to accurately predict hand pose when the fingers are occluded. 
}

\begin{figure}
    \centering
    \includegraphics[width=.90\linewidth]{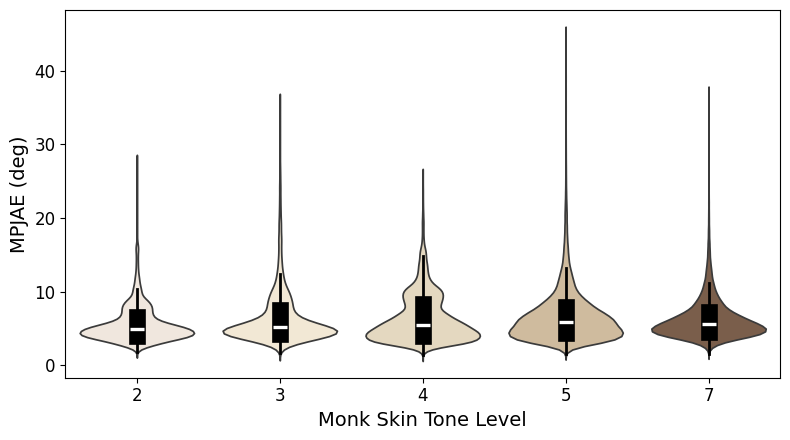}
    \caption{Distribution of MPJAE compared to Monk skin tone level. Violin plots are colored by their corresponding skin tone.}
    \label{fig:monk-analysis}
    \Description{
    Violin plots of MPJAE over Monk Skin Tone Level. Violin plots all have similar distributions, with the median hovering around MPJAE=7 and interquartile ranges from 0-10. 
    }
\end{figure}

\textbf{Performance across skin tones. }
Using the predicted poses from all LOSO experiments, we then conduct a one-way analysis of variance (ANOVA) \cite{gillardOneWayAnalysisVariance2020} to analyze the effect of skin tone on performance. We find a statistically significant effect ($p=4.07\times10^{-205}$) but small effect size ($\eta^2=0.0055$) indicating that skin tone explained less than 1\% of the data variance and does not meaningfully affect hand pose prediction accuracy. These results indicate that our approach using dorsal deformations is inclusive across people with varying skin tones.

% \usepackage{multirow}

% \usepackage{multirow}
% \usepackage{graphicx}
\begin{table*}[]
\caption{Average results of leave-one-subject-out experiments across all participants for each gestures.  ($\text{Mean}_{\text{participants}} \pm \text{SD}_\text{participants}$).}
\label{tab:performance-gesture-results}
% \resizebox{.8\linewidth}{!}{%
\changed{
\small
\begin{tabular}{l|l|ccc}
Gesture & Metric & HaMeR \cite{pavlakosReconstructingHands3D2023} & HandOccNet \cite{parkHandOccNetOcclusionRobust3D2022} & Ours \\ \hline
\multirow{2}{*}{Tap} & MPJAE ($\degree$) & \phantom{1}$ 5.45 \pm 0.66 $ & \phantom{1}$ 9.46 \pm 1.14 $ & \phantom{1}$ \mathbf{5.43 \pm 0.89} $ \\
 & PA-MPJPE (mm) & \phantom{1}$ \mathbf{6.33 \pm 0.65}$ & $ 12.72 \pm 1.56$ & \phantom{1}$ 6.73 \pm 0.87$ \\ \hline
\multirow{2}{*}{Pinch} & MPJAE ($\degree$) & \phantom{1}$ 5.90 \pm 1.02 $ & $ 10.65 \pm 0.88 $ & \phantom{1}$ \mathbf{5.75 \pm 0.98} $ \\
 & PA-MPJPE (mm) & \phantom{1}$ \mathbf{6.33 \pm 0.81} $ & $ 15.08 \pm 1.05$ & \phantom{1}$ 6.74 \pm 0.98$ \\ \hline
\multirow{2}{*}{Curl} & MPJAE ($\degree$) & \phantom{1}$ 7.50 \pm 1.12 $ & $ 12.24 \pm 1.90 $ & \phantom{1}$ \mathbf{6.81 \pm 1.11} $ \\
 & PA-MPJPE (mm) & \phantom{1}$ \mathbf{7.43 \pm 0.75}$ & $ 14.26 \pm 0.98$ & \phantom{1}$ 7.45 \pm 0.92$ \\ \hline
\multirow{2}{*}{Bear Claw} & MPJAE ($\degree$) & \phantom{1}$ 9.91 \pm 1.85 $ & $ 16.53 \pm 2.60 $ & \phantom{1}$ \mathbf{8.09 \pm 2.26} $ \\
 & PA-MPJPE (mm) & \phantom{1}$ 7.90 \pm 0.85$ & $ 16.11 \pm 1.74$ & \phantom{1}$ \mathbf{7.36 \pm 1.37}$ \\ \hline
\multirow{2}{*}{Fist} & MPJAE ($\degree$) & $ 13.60 \pm 3.00 $ & $ 19.70 \pm 1.50 $ & \phantom{1}$\mathbf{ 9.94 \pm 2.95} $ \\
 & PA-MPJPE (mm) & $ 11.53 \pm 2.41$ & $ 17.13 \pm 1.14$ & \phantom{1}$ \mathbf{9.83 \pm 2.49}$ \\ \hline
\multirow{2}{*}{Fan} & MPJAE ($\degree$) & \phantom{1}$ \mathbf{6.51 \pm 1.08} $ & $ 10.26 \pm 1.85 $ & \phantom{1}$ 6.58 \pm 1.16 $ \\
 & PA-MPJPE (mm) & \phantom{1}$ \mathbf{7.53 \pm 1.37} $ & $ 14.47 \pm 2.74$ & \phantom{1}$ 8.22 \pm 1.49$ \\ \hline
\multirow{2}{*}{Free} & MPJAE ($\degree$) & \phantom{1}$ \mathbf{7.86 \pm 1.14} $ & $ 12.38 \pm 1.48 $ & \phantom{1}$ 8.16 \pm 1.33 $ \\
 & PA-MPJPE (mm) & \phantom{1}$ \mathbf{8.69 \pm 1.52}$ & $ 16.16 \pm 1.87$ & \phantom{1}$ 9.98 \pm 1.72$
\end{tabular}
% }
}
\end{table*}

\begin{table}[]
\caption{Average performance of \systemname across different image sizes as input. Values are averaged across three different participants of varying skin tones (P1, P5, P10). $384\times384$ is the default setting of \systemname. \changed{Inference time was measured on a single NVidia RTX A5500}. ($\text{Mean}_{\text{participants}} \pm \text{SD}_\text{participants}$).}
\label{tab:img-size-ablation}
% \resizebox{.6\linewidth}{!}{%
\small
\begin{tabular}{l|ccc}
Image Size & MPJAE ($\degree$) & PA-MPJPE (mm) & Inference Time (ms) \\ \hline
$128\times128$ & \phantom{1}$7.57 \pm 7.82$ & \phantom{1}$9.04 \pm 7.56 $ & \phantom{1}$\mathbf{5.20 \pm 0.33}$ \\
$256\times256$ & \phantom{1}$7.00 \pm 7.34$ & \phantom{1}$\mathbf{7.80 \pm 6.95} $ & \phantom{1}$9.61\pm 0.26$ \\
$384\times384$ & \phantom{1}$\mathbf{6.80 \pm 7.16}$ & \phantom{1}$7.93 \pm 6.89  $ & $20.95 \pm 0.21$
\end{tabular}
% }
\end{table}

\changed{
\textbf{Performance across gestures.}
When analyzing by gesture type, we find our system matches HaMeR's performance for multiple gestures and outperforms HandOccNet for all gestures, indicating that \systemname can enable more fine-grained and precise gesture recognition and interaction schemes that leverage partial gestures like small or large taps under occlusive scenarios when the fingers are not visible. When analyzing the performance of our system across different gestures, we find that our system performs worse for ``fan'' pose reconstruction (PA-MPJPE: Ours: $8.22\pm1.49$mm, HaMeR: $7.53\pm1.37$mm), and matches performance for ``tap'', ``pinch'', and ``curl''. We believe this can be attributed to the fact that these gestures are often less self-occluded when the camera is pointing downwards as in our experimental setup. This is further supported by strong performance for ``fist'' and ``bear claw'' (MPJAE: $9.94\pm 2.95\degree$ and $8.09\pm2.26\degree$ respectively), outperforming HaMeR by over 3 degrees. (\Cref{tab:performance-gesture-results}). This can be partially attributed to the large changes in the dorsal feature around the knuckles and tendons during these gestures which are easier for our system to pick up. However, towards the apex of these gestures, we notice that skin deformations are largely contained to dorsal features stretching across the hand. These fine-grained skin feature translations are hard to capture and often missed in our system, which reduces images to $16\times16$ pixel feature patches before encoding. 
}

\subsection{Ablation Study}

\begin{table*}[]
\caption{\changed{Average performance of \systemname across different delta stream DINOv3 backbones. Values are averaged across three different participants of varying skin tones (P1, P5, P10). ViT-L is the default setting of \systemname. \changed{Inference time was measured on a single Nvidia RTX A5500}. ($\text{Mean}_{\text{participants}} \pm \text{SD}_\text{participants}$).}}
\label{tab:backbone-ablation}
\small
\changed{
\begin{tabular}{lc|c|ccc}
Backbone & Parameters & Delta Stream & MPJAE ($\degree$) & PA-MPJPE (mm) & Inference Time (ms) \\ \hline
\multirow{2}{*}{ViT-S+} & \multirow{2}{*}{29M} & \ding{55} & $24.31 \pm 29.45 $ & $19.00 \pm 15.51$ & \phantom{1}$\mathbf{3.97}$ \\
 &  & \ding{51} & $7.13 \pm 7.45$ & $8.28 \pm 6.98$ & \phantom{1}4.54 \\ \hline
\multirow{2}{*}{ViT-B} & \multirow{2}{*}{86M} & \ding{55} & $7.06 \pm 7.43$ & $8.30 \pm 7.02$ & \phantom{1}7.54 \\
 &  & \ding{51} & $6.98 \pm 7.37$ & $8.15 \pm 7.16$ & \phantom{1}7.97 \\ \hline
\multirow{2}{*}{ViT-L} & \multirow{2}{*}{300M} & \ding{55} & $\mathbf{6.79 \pm 7.29}$ & $\mathbf{7.86 \pm 6.93}$ & 20.33 \\
 &  & \ding{51} & $6.80 \pm 7.16$ & $7.93 \pm 6.89$ & 20.95
\end{tabular}
}
\end{table*}

To better understand what \systemname is using to predict hand pose, we conducted two ablation studies on the input image size and DINOv3 backbone. We follow the evaluation procedures described in \Cref{sec:perf-study} and test each configuration on three subjects with varying skin tones (P1: Monk skin tone level 5, P5: Monk skin tone level 2, P10: Monk skin tone level 7) to reduce training time.

\textbf{Image size.}
We found that our base image input size of $384\times384$ was enough to preserve the clarity of the dorsal area of the hand in the 4K camera capture. By downscaling these images, our system is restricted to more low-frequency features of the hand like large changes in shadows and tendons. Notably, existing pose estimation methods often resize a segmentation of the whole hand to $256\times256$ or even smaller \cite{pavlakosReconstructingHands3D2023, Zhang2020MediaPipe, parkHandOccNetOcclusionRobust3D2022}, destroying fine skin features. As shown in \Cref{tab:img-size-ablation}, we observe performance drop-offs in pose reconstruction as image size decreases, indicating that fine skin features, such as wrinkles, contribute to the overall performance of the model. Performance for fine movements, such as small taps and pinches, is especially affected due to the fact that they mostly only induce small skin translations. By lowering the image size, we also find significant improvements to inference time, potentially enabling real-time pose estimation through dorsal features. We also note that training at lower image sizes led to more unstable training loss.

\textbf{Backbone.}
We also evaluated different backbone sizes and delta stream configurations for DINOv3 to assess how much model capacity is required to capture dorsal features in dense visual embeddings and the usefulness of the proposed delta stream. The different backbone sizes are distillations of DINOv3 which can also be used as a measure of backbone complexity where ViT-S+ can be implemented into small mobile computing systems. When analyzing the full delta-stream network, both MPJAE and PA-MPJPE increased only marginally with much smaller backbones indicating that dorsal deformations can be effectively encoded without high-dimensional models (\Cref{tab:backbone-ablation}). At the same time, inference time dropped substantially, from 20.95 ms with ViT-L to 4.54 ms with  ViT-S+. These findings suggest that \systemname can be deployed on mobile devices with minimal performance loss. 
\changed{Furthermore, while the delta stream had little impact on the performance of the ViT-L DINOv3 backbone, the removal of delta features in the pipeline had a significant negative on the performance of smaller backbones. We attribute this to the fact that as features become simpler at lower backbones, the delta features become more useful to specifically isolate dorsal changes. This hypothesis is further supported as model collapse or nonconvergence was nearly unavoidable for the ViT-B and ViT-S+ backbones. Our results indicate that delta features can be a useful paradigm for future small mobile computing systems which can only deploy small ViT models due to hardware constraints.}

\section{Sensing Tap, Pinch, and Click}
We follow common HCI techniques to analyze how \systemname can improve gesture accuracy in different self-occluded scenarios. For all subsequent analyses using pose, we follow the same procedures as described in \Cref{sec:perf-study} but report values across all testing sets for simplicity. We focus on HaMeR as a baseline, as HandOccNet showed limited effectiveness in our setting. Finally, we analyze the cross-applicability of dorsal deformations by using our approach to detect isometric ``force clicks'' with no discernible hand motion.

\subsection{\changed{Self-Occluded Tap Detection}}
\label{sec:tap-application}

\begin{table}[]
\caption{\changed{Results of self-occluded tap application experiments across all 12 LOSO experiments}. RMSE refers to each joint's X-axis root mean squared angular error. }
\label{tab:tap-angle-application-results}
\small
\changed{
\begin{tabular}{l|cc}
Metric & HaMeR \cite{pavlakosReconstructingHands3D2023} & Ours \\ \hline
Index RMSE ($\degree$) & 31.15 & \textbf{26.23} \\
Middle RMSE ($\degree$)  & 26.02 & \textbf{16.07} \\
Pinky RMSE ($\degree$) & \textbf{11.29} & 15.89
\end{tabular}
}
\end{table}

\changed{
We first focus on taps and use the common heuristic of X-axis Euler rotation (rotation axis perpendicular to the finger and palm normal) of the metacarpal joint as a measurement of tap completion, which can be used as a continuous value for fine-grained control or a binary ``tap'' / ``no tap'' through a simple threshold. We analyze only taps where the finger is over 50\% occluded to measure partial self-occlusion performance in HaMeR. We do not evaluate the ring finger due to participant discomfort in the data collection process. When evaluating across our collected tap data, we find that \systemname significantly reduces the angular RMSE of both the index and middle finger. We attribute this difference to the fact that when tapping the index and middle finger, the entire dorsal region tends to stretch across the knuckle more amplifying our signal. This is further supported as \systemname performs worse on pinky tap detection (+4.60\degree) which we attribute to the fact that the pinky induces a far smaller dorsal skin change when tapped. Thus, \systemname demonstrates better capabilities for fine-grain control in small gestures like taps when the fingers are not totally visible and dorsal deformations are easily isolated like in the index finger, with more limited applications in the pinky which appears robust for coarse event detection but not continuous precision control. We believe that this can generalize to other interactions like surface taps by reducing angular error in microgesture scenarios and even non-occluded cases by using the dorsal region as another prior to predict depth.
}

\subsection{\changed{Self-Occluded Pinch Detection}}
\label{sec:pinch-application}

\begin{table}[]
\caption{\changed{Results of self-occluded pinch application experiments across all 12 LOSO experiments. Dist. RMSE refers to the fingertip-to-thumb distance root mean squared error.}}
\label{tab:pinch-application-results}
\small
\changed{
\begin{tabular}{l|cc}
Metric & HaMeR \cite{pavlakosReconstructingHands3D2023} & Ours \\ \hline
Index Dist. RMSE (mm) & 25.46 & \textbf{16.81} \\
Middle Dist. RMSE (mm) & \textbf{21.23} & 26.33 \\
Ring Dist. RMSE (mm) & \textbf{20.08} & 23.17
\end{tabular}
}
\end{table}

\changed{
Following our approach in \Cref{sec:tap-application}, we use the common heuristic of finger tip to thumb tip distance as a measurement of pinch completion, which can be used as a continuous value for fine-grained control or binary classification, and analyze at least 50\% occluded pinch scenarios. When comparing against HaMeR on our collected pinch data (\Cref{tab:pinch-application-results}), we find that \systemname significantly outperforms HaMeR in index pinches (Dist. RMSE: $-8.65$mm) but performs worse for both middle and ring finger pinches (Dist. RMSE: +$5.10$mm and +$3.09$mm respectively). We believe this is caused by how people generally pinch, where middle finger and ring finger pinches often induce small dorsal region changes as the thumb moves towards the finger. This phenomenon is reduced in the index finger which we found moves more during pinching. 
}

\subsection{Isometric Click Detection}

\begin{figure*}
    \centering
    \includegraphics[width=0.98\linewidth]{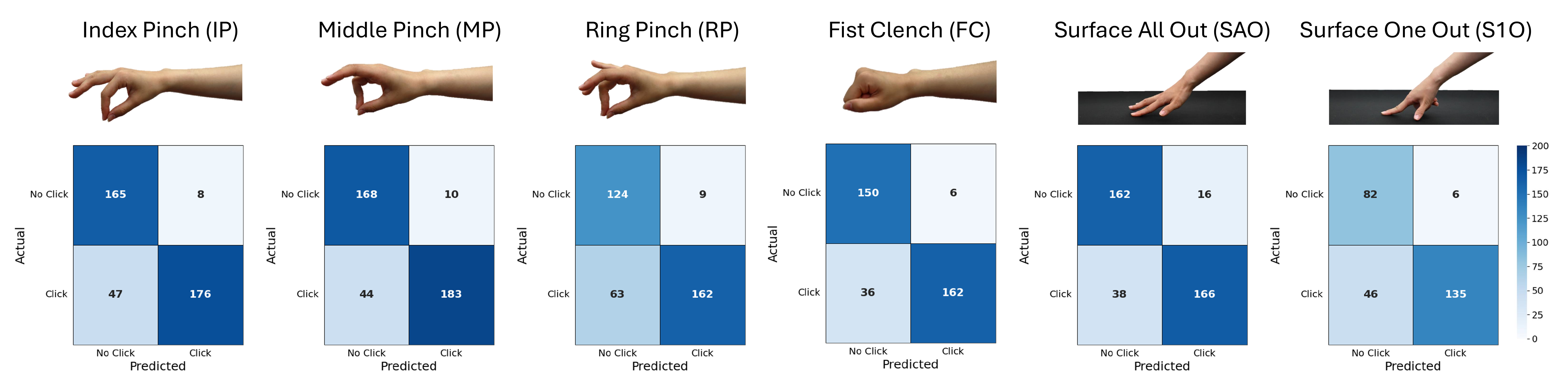}
    \caption{The six static gestures for force data collection and confusion matrices for each gesture type. Values are reported per click rather than per frame following common HCI systems.}
    \label{fig:force-data}
    \Description{
    Six images of different static click gestures and their corresponding confusion matrices. The gestures are as follows: index pinch (IP), middle pinch (MP), ring pinch (RP), fist clench (FC), surface all out (SAO) (all fingers out pressing a surface with the index finger), surface one out (S1O) (all fingers curled in with the index finger out pressing a surface). Each confusion matrix is a 2x2 with no click and click. All confusion matrices show a majority of correct predictions, with a large amount of false positives. Surface one out had the worst matrix, with 46 of 181 actual clicks predicted as no click.
    }   
\end{figure*}

Dorsal skin deformations are also induced by the hand when exerting a force, even in static positions. During our data collection process, we also collected 30 second trials of force data from 4 static in-air gestures (index finger pinch, middle finger pinch, ring finger pinch, fist clench) and two surface gestures (surface index finger tap with all fingers out, surface index finger tap with one finger out) from 11 participants (P2-P12) using a force-sensing resistor (FSR) synchronized with the camera capture using the serial output from an Arduino. Gestures are shown in \Cref{fig:force-data}. We normalized all FSR readings by the max force exerted in the trial to control for a participant's hand strength and defined a click as a force reading peak above 20\% of the trial's max force reading. In total, we collected 41,894 frames of data. 

We then trained a simplified version of \systemname using only the delta features ($F_t-F_0$, $\cos (F_t, F_0)(x,y)$) from a totally frozen DINOv3 featurizer as input into a classification head to analyze the cross-applicability of our feature difference approach in different interaction applications. Our model was trained under cross entropy loss. We trained three models under an 80/10/10 data split due to our limited data size and report the model with the best performing validation metrics. While our model predicts click per frame, we follow HCI conventions and report all metrics per click where each click is labeled through majority vote to better present how systems like ours are often implemented. Predictions are reduced to a binary classification per click as detecting the specific gesture is trivial through pose estimation as discussed in \Cref{sec:perf-study}. More information on our model can be found in our codebase. 

\textbf{Results.}
\Cref{fig:force-data} shows the confusion matrix with the heatmap of all 6 gestures. Accuracy across our entire testing set was 0.85 with a weighted precision of 0.86, weighted recall of 0.85, and weighted F1-score of 0.85. We found noticeably more false negatives than positives. We believe this is partially caused by small changes in the overall hand position, which can stretch the skin more taut and obscure the small dorsal deformations induced by clicks. 

Our overall accuracy across all in-air click interactions was 0.85 with the best performing gesture being fist clench (FC) (0.88) aligning with intuition that FC would induce the most dorsal change given the soft palm and amount of force a user can apply. Of the pinch gestures, middle pinch (MP) performed the best with an accuracy of 0.87 while ring pinch (RP) performed the worst with an accuracy of 0.80. We attribute this performance decrease to how unnatural RP is for users to do, causing their hand to tense up more and obscuring small dorsal deformations in the process.

Our two surface clicks performed similarly with an overall accuracy of 0.84. Notably, Surface all out (SAO) noticeably outperformed surface one out (S1O) (0.86 and 0.81 respectively). Similarly to the RP, since the user's fingers are pulled in for S1O, dorsal skin is pulled more taut which dampens the dorsal deformations caused by clicks. Furthermore, we see that in SAO, the tendons under the skin are more relaxed, inducing a larger change when clicking compared to S1O.

Our results indicate that our approach using purely dorsal changes can generalize beyond pose, enabling a new dimension of XR interactions that can fuse with traditional pose-based gestures. We believe that dorsal changes should play a key role in future click and static interaction detection systems, similar to how they fuse with the original image features in our pose estimation architecture (\Cref{fig:system-architecture}).
\section{Discussion}
We discuss key implications of our findings, limitations, and opportunities for future work, specifically through existing egocentric systems.

\textbf{Addressing self-occlusion.}
Through our motivating study in \Cref{sec:motivation}, we confirm the impact of self-occlusion on pose estimation performance demonstrated in prior works \cite{parkHandOccNetOcclusionRobust3D2022, liCHORDCategorylevelHandheld2023}. Our analysis shows that over 20\% of egocentric camera captures are affected by self-occlusion, which can increase pose estimation angular error by over $15$ degrees in SOTA models (\Cref{fig:baseline-perf-occ}). This severely limits existing systems, forcing users to keep their fingers in view of the egocentric camera to detect a specific gesture. By utilizing the more visible dorsal area, we are able to mitigate these effects and enable a more expansive interaction area without the need for new sensors beyond an egocentric camera. Future work should continue to focus on methods to address the impacts of self-occlusion, fusing in temporal pose features and semantic gestural prediction to create a more seamless interaction experience. 

% enabling more fine grain gestures and control in a new dimension which previously wasnt super possible (fine grain hand gestures and clicking) invariant of skin color
\textbf{Enabling fine-grained control of interactions.}
As shown in \Cref{sec:motivation}, current SOTA vision-based pose estimation systems are still unreliable in everyday egocentric video capture, especially when it comes to small and precise movements and gestures. This can be partially attributed to a lack of depth sensing, making it difficult to discern the precise angle of a finger. Through our evaluations and applications, \systemname demonstrated improved capabilities for capturing small shifts in gestures, enabling new interactions like static clicks, partial pinch, and partial tap that were previously not recognizable in self-occluded scenarios. Furthermore, we believe our system can also enable new micro-gestures, which have already demonstrated their utility in prior HCI research \cite{kinSTMGMachineLearning2024, liuHoloscopic3DMicroGesture2018, luHCMGHumanCapacitanceBased2024, sangMicroHandGesture2018}, without the need for new sensors or more invasive apparatuses. Further work is needed to expand our dataset to microgestures that may be fully visible in the camera, but are difficult for existing pose estimation models to detect. We note that our technical approach to capturing dense skin features is still relatively simple as we only use the pretrained DINOv3 model weights. Prior work has demonstrated that the DINO architecture can be significantly improved when pretraining on domain-specific data over a fine-tune \cite{simeoniDINOv32025}, potentially enabling the use of smaller architectures and better accuracy. Thus future work should also focus on computational techniques to best capture fine-grain skin features.

\textbf{Generalizability.}
While our system outperformed SOTA baselines when tested on our lab dataset, more work is needed to understand how our approach can generalize to in-the-wild scenarios. Our LOSO experiments and skin tone study results indicate that leveraging skin deformations can generalize to out-of-distribution subjects and serve as a reliable data source even without model personalization or fine-tuning. Future work should aim to conduct a study with a larger population with a wider range of skin, skin age, hair characteristics, lighting conditions, and blemishes to better understand how the utility of dorsal features can change depending on the population and setting at hand.

Secondly, we acknowledge that our approach leveraging purely dorsal features for pose estimation is only applicable for a portion of in-the-wild scenarios where the dorsum of the hand is visible. \changed{Since our analysis was on a constrained dataset, more work is needed to analyze the generalizability of \systemname in real-time and unconstrained scenarios.} We believe our approach integrating more general pose estimation models presents a blueprint for how future systems should be integrated, leveraging the best possible signal---hand silhouette or dorsal skin---when available. In such situations, trivial estimations like hand orientation can serve as powerful confidence scores to dictate how useful a signal is.

\textbf{Deployability.}
We found through our ablation tests (\Cref{tab:backbone-ablation}) that small lightweight backbones like ViT-S+ could reasonably provide usable performance for far less compute and shorter inference time. Future work should continue to investigate this idea, offloading detail capture to high-resolution sensors which can then be processed faster and more efficiently on wearable devices like lightweight AR glasses and more which previously could not support accurate single-camera hand tracking.

We also note that \systemname requires a high-resolution camera feed to maximize performance and capture dorsal features during motion blur induced by fast movements of the hand and head, requiring power and processing. However, high quality egocentric head-worn cameras are already starting to be integrated into headsets \cite{AppleVisionPro, Pico4Review2022, storeDiscoverRayBanMeta}. We acknowledge that using this hardware for hand tracking will likely increase power draw and computational complexity. Nonetheless, we believe \systemname can still make a tangible impact in existing systems given that it enables the utilization of the large amounts of monocular egocentric data are currently readily available. 

\systemname is also a far smaller model than HaMeR with similar performance across all scenarios, self-occluded or not, allowing for deployment onto more hardware constrained devices that may not be able to to support and run a ViT-H based backbone.

Finally, we note that our approach can still be applied to low resolution images for low-cost monocular sensing systems as demonstrated in our ablation tests on both image size (\Cref{tab:img-size-ablation}) and backbone size (\Cref{tab:backbone-ablation}). The camera can also be mounted closer to the hand (\eg, on the chest) to improve the resolution of dorsal features without changing camera quality.

% new dimension of high frequency data which was previously ignored in DL and CV
\textbf{Leveraging high-frequency features.}
While previous pose estimation methods were limited to low-resolution images of hands due to computational or architectural constraints, recent advances in dense featurization in computer vision have now enabled us to integrate finer features that we previously could not capture and use. \systemname demonstrates this very idea by using purely skin deformations, a high frequency signal, which can match SOTA performance without ever seeing the fingers. Our approach follows the findings of recent works in vision transformers which show that by lowering the patch size and analyzing more of the fine-grained features in an image, overall model performance improves \cite{wangScalingLawsPatchification2025a, liuMSPEMultiScalePatch2024, beyerFlexiViTOneModel2023a, nguyenImageWorthMore2024}. Thus, we believe that future works should continue to integrate fine-grained signals to enhance existing models, like in the case of \systemname, and enable new interactions. In the meantime, more work is needed to identify how new sensors should adapt to capture this new frontier of signal. For instance, head-mounted cameras must capture at a higher resolution and refresh rate to properly capture dorsal features or integrate polarization cameras to isolate skin wrinkles.

\changed{
\textbf{Beyond hand pose estimation.}
\systemname uses changes in dorsal skin induced by the complex tendon activations and tension in the skin to predict pose. These same features can extend beyond hand sensing, enabling grasp and shape detection in held objects as demonstrated in Zhai et al. \cite{zhaiInfluenceGraspingPostures2023}. Beyond just grasp, the same features that were used for isometric click detection can also be used to sense the hardness of held objects, enabling both hand and object detection. Dorsal skin has also historically been used in medical sensing, serving as a strong signal for aging \cite{gaoSkinTextureParameters2011} and dehydration \cite{goehringMeasuresSkinTurgor2022}, as well as general tendon usage. By analyzing specifically dorsal activation, we can also build models of biomechanical activation which can enable generic wellness and hand fatigue tracking in long-form egocentric use cases.
}

% testing in real world scenarios and architectural changes
\section{Conclusion}

In this paper, we introduce \systemname, a deep learning system for 3D pose estimation from egocentric perspectives by detecting deformations in the dorsum of the hand. Our system is the first to leverage dorsal features atemporally without any physical localization, allowing for more flexible camera positions and robust systems that can easily integrate with existing 3D pose estimation models without the need to see full hand silhouettes. 

We first demonstrate that over 20\% of all egocentric videos demonstrate significant amounts of hand self-occlusion which can substantially reduce the performance of existing SOTA pose estimation models (HaMeR: $m=-8.42\degree$). We then collect a dataset of high-resolution hand captures from an egocentric perspective to evaluate our approach. Our results show that \systemname outperforms existing SOTA baselines using purely dorsal features with a mean per joint angular error of 6.41$\degree$. Our system is generalizable, with results showing \systemname is less affected by self-occlusion ($m=-4.59\degree$) than SOTA baselines and skin-tone agnostic ($\eta^2=0.0055$). We then demonstrate how \systemname enhances the usability of XR interactions by improving pose prediction and recognition for two key gestures: pinch and tap. Finally, we show that our approach isolating skin deformations is applicable across non-pose-based gestures like in-air and surface clicks (accuracy: 0.85 and 0.84 respectively).

While we believe \systemname is only a step towards a more embodied and seamless interaction experience in XR, we hope our method and results provide insights into a new approach to sensing human signals through dense features previously ignored.

% \acknowledgements{}

\begingroup
\sloppy
\bibliographystyle{ACM-Reference-Format}
\bibliography{ref.bib}
\endgroup
\clearpage
\appendix
\newpage
\appendix
\renewcommand{\thefigure}{\thesection.\arabic{figure}}

\renewcommand{\thetable}{\thesection.\arabic{table}}

\setcounter{figure}{0}
\setcounter{table}{0}

\section{Appendix}

% \usepackage{booktabs}
\begin{table}[H]
\caption{Demographics and hand information of 12 participants (P1-P12) in the data collection. Monk level indicates the participants Monk Skin Color Level from a scale of 1-10. $\frac{\text{Circumference}}{\text{Length}}$ is the ratio of hand circumference to length where higher is a wider hand while lower is a longer hand. Hairiness indicates the participant's self-reported hand hairiness on a scale from 1-5 where 1 is no hair and 5 is extremely hairy.}
\label{tab:participant-info}
% \resizebox{.4\linewidth}{!}{%
\small
\begin{tabular}{@{}cccccc@{}}
\toprule
ID & Age & Gender & Monk Level & $\frac{\text{Circumference}}{\text{Length}}$ & Hairiness \\ \midrule
P1 & 23 & M & 5 & 1.05 & 2 \\
P2 & 23 & F & 4 & 0.94 & 1 \\
P3 & 20 & F & 5 & 1.06 & 1 \\
P4 & 26 & F & 3 & 1.12 & 1 \\
P5 & 29 & F & 2 & 1.00 & 1 \\
P6 & 23 & F & 3 & 1.13 & 1 \\
P7 & 24 & F & 5 & 1.03 & 1 \\
P8 & 25 & M & 7 & 1.03 & 2 \\
P9 & 18 & M & 5 & 1.14 & 3 \\
P10 & 26 & M & 7 & 0.97 & 4 \\
P11 & 23 & M & 4 & 1.09 & 1 \\
P12 & 20 & M & 7 & 1.05 & 1 \\ \bottomrule
\end{tabular}
% }
\end{table}

\begin{figure}[H]
    \centering
    % First subfigure
    \includegraphics[width=.45\linewidth]{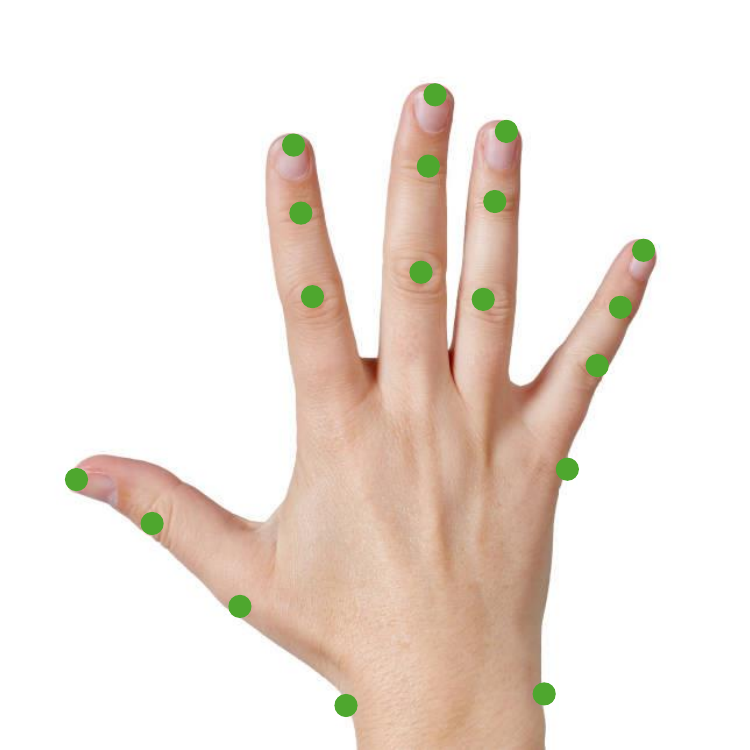}
    \caption{Our custom marker definition for motion capture during data collection.}
    \label{fig:vicon-marker-def}
    \Description{
    Visualization of a hand with markers indicating the motion capture marker position. All markers are on top of the fingers and on the side of the hand. Notably, no markers are placed on the dorsal area.
    }
\end{figure}
  
\end{document}